\documentclass[lettersize,journal]{IEEEtran}
\usepackage{amsmath,amsfonts}
\usepackage{algorithmic}
\usepackage{algorithm}
\usepackage{array}
\usepackage[caption=false,font=normalsize,labelfont=sf,textfont=sf]{subfig}
\usepackage{textcomp}
\usepackage{stfloats}
\usepackage{url}
\usepackage{verbatim}
\usepackage{graphicx}
\usepackage{cite}
\usepackage[colorlinks=true, linkcolor=blue, citecolor=blue, urlcolor=blue]{hyperref}

\usepackage{booktabs}

\usepackage{orcidlink}

\usepackage{url}

\usepackage{amsfonts, bm}

\usepackage{multirow} 
\usepackage{makecell} 

\usepackage{colortbl}

\definecolor{unit01green}{RGB}{82,208,83}

\definecolor{unit02red}{RGB}{211,41,15}
\newcommand{\red}[1]{\textcolor{unit02red}{#1}}

\definecolor{unit02blue}{RGB}{53,49,255}
\newcommand{\blue}[1]{\textcolor{unit02blue}{#1}}

\definecolor{lightblue}{rgb}{0.68, 0.85, 0.9}

\definecolor{grey}{rgb}{0.9, 0.9, 0.9}
\newcommand{\ccol}{\cellcolor{grey}}

\newcommand{\highlight}[1]{{\cellcolor[rgb]{0.925,0.957,1}}{#1}}

\DeclareMathOperator*{\argmax}{arg\,max}

\begin{document}

\title{MADS: Multi-Attribute Document Supervision for Zero-Shot Image Classification}

\author{
Xiangyan Qu, 
Jing Yu$^{*}$,~\IEEEmembership{Member,~IEEE,}
Jiamin Zhuang, 
Gaopeng Gou, 
Gang Xiong,
Qi Wu,~\IEEEmembership{Member,~IEEE,}
\thanks{This work was supported by the Central Guidance for Local Special Project (Grant No. Z231100005923044).}
\thanks{Xiangyan Qu, Jiamin Zhuang, Gaopeng Gou, Gang Xiong are with the Institute of Information Engineering, Chinese Academy of Sciences, China, and the School of Cyber Security, University of Chinese Academy of Sciences, China. (e-mail: quxiangyan@iie.ac.cn; zhuangjiamin@iie.ac.cn; gougaopeng@iie.ac.cn; xionggang@iie.ac.cn)}
\thanks{Jing Yu is with the School of Information Engineering, Minzu University of China, Beijing, China. (e-mail: jing.yu@muc.edu.cn)}
\thanks{Qi Wu is with the Australia Centre for Robotic Vision (ACRV), the University of Adelaide. (e-mail: qi.wu01@adelaide.edu.au)}
\thanks{Corresponding author: Jing Yu (e-mail: jing.yu@muc.edu.cn)}
}

\markboth{Journal of \LaTeX\ Class Files,~Vol.~14, No.~8, August~2021}%
{Shell \MakeLowercase{\textit{et al.}}: A Sample Article Using IEEEtran.cls for IEEE Journals}


\maketitle

\begin{abstract}
Zero-shot learning (ZSL) aims to train a model on seen classes and recognize unseen classes by knowledge transfer through shared auxiliary information. Recent studies reveal that documents from encyclopedias provide helpful auxiliary information. However, existing methods align noisy documents, entangled in visual and non-visual descriptions, with image regions, yet solely depend on implicit learning. These models fail to filter non-visual noise reliably and incorrectly align non-visual words to image regions, which is harmful to knowledge transfer. In this work, we propose a novel multi-attribute document supervision framework to remove noises at both document collection and model learning stages. With the help of large language models, we introduce a novel prompt algorithm that automatically removes non-visual descriptions and enriches less-described documents in multiple attribute views. Our proposed model, MADS, extracts multi-view transferable knowledge with information decoupling and semantic interactions for semantic alignment at local and global levels. Besides, we introduce a model-agnostic focus loss to explicitly enhance attention to visually discriminative information during training, also improving existing methods without additional parameters. With comparable computation costs, MADS consistently outperforms the SOTA by 7.2\% and 8.2\% on average in three benchmarks for document-based ZSL and GZSL settings, respectively. Moreover, we qualitatively offer interpretable predictions from multiple attribute views. 
\end{abstract}



\section{Introduction}
\label{sec:intro}

\IEEEPARstart{Z}{ero-shot} learning (ZSL) \cite{ZSL_invent1, ZSL_invent2, ZSL_invent3} aims to train a model solely on a set of seen classes to identify unseen classes whose categories with no labeled samples available.
Recently, ZSL has garnered promising interest as a recognition challenge miming the human cognitive process without enormous manually annotated data \cite{ZSL_invent1}.
Humans possess an impressive ability to recognize images of unseen classes through prior knowledge of the seen domain.
Following this idea, leveraging the auxiliary information to transfer knowledge from seen to unseen classes becomes a core problem of ZSL.

Common auxiliary information includes attributes~\cite{ZSL_invent3, AWA1_and_attribute_provide}, word embeddings~\cite{word_embed_NIPS_2013_Richard_Socher, Word_embed_as_AUX_1}, and documents~\cite{document_as_AUX_1, Document-based_ICCV_2015_Lei_Jimmy_Ba}.
Most work \cite{CVPR_2016_Yongqin_Xian, CVPR_2022_MSDN, CVPR_2023_PSVMA, GZSL_ICML_2023, CVPR_2024_GEN_1, semantic_align_ZSLViT} uses human-annotated attributes as auxiliary information, which defines seen and unseen classes with the same attributes. 
These methods transfer knowledge by aligning attributes with corresponding image regions in the seen domains.
However, annotating attributes is labor-intensive and time-consuming ~\cite{Attribute_drawback_CVPR_2013, Attribute_drawback_ECCV_2018}. 
Moreover, each novel category requires distinct attribute values and even new attributes, which is hard to scale in real-world scenarios~\cite{Attribute_drawback_and_document_arxiv_2021}.
In contrast, other work \cite{word_embed_NIPS_2013_Devise, SJE, word_embed_and_graph_CVPR_2019_Michael_Kampffmeyer, VGSE} utilizes word embedding of category names instead of attributes. 
They build the relationship between word embeddings of seen and unseen classes within the common language space to transfer knowledge.
However, the class name offers minimal discriminative information and sensitivity to linguistic issues\cite{SJE, Attribute_drawback_and_document_arxiv_2021}, limiting the performance.
Recent work \cite{Document-based_NAACL_2021_Jihyung_Kil, Document-based_CVPR_2017_Mohamed, GAZSL, CIZSL, LsrGAN, I2DFormer, I2MVFormer} leverages encyclopedic documents, \textit{e.g.}, Wikipedia, as auxiliary information (denoted as document-based ZSL) to address above issues in attributes and word embeddings.
These documents contain rich semantic knowledge from human experts and are easily accessible online.


\IEEEpubidadjcol

{
\begin{figure*}[!t]
\begin{center}
\centerline{\includegraphics[width=0.75\linewidth]{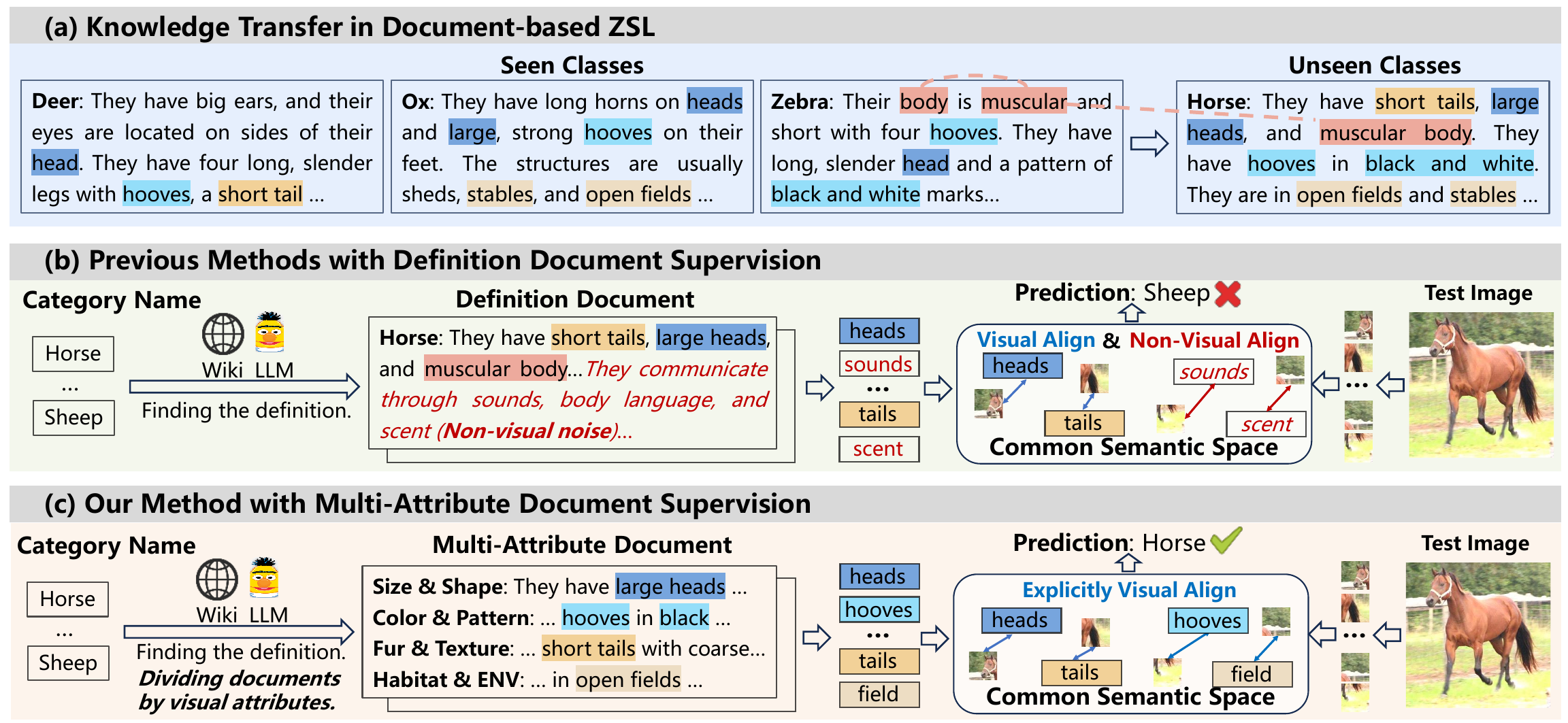}}
\vskip -0.05in
\caption{
(a) In document-based ZSL, visual words bridge knowledge transfer from seen to unseen classes (shown in the same color).
(b) Previous methods align noisy documents with image regions, which is detrimental to knowledge transfer.
(c) In contrast, we automatically remove non-visual noise before model training to obtain multi-attribute documents and then explicitly align visual words of each attribute view with salient image regions.
}
\label{fig: motivation}
\end{center}
\vskip -0.15in
\end{figure*}
}

As shown in Figure \ref{fig: motivation}(a), visual words (adjectives and nouns that describe visual details) are shared knowledge between seen and unseen classes. For example, descriptions about ``\texttt{short tails}'' and ``\texttt{muscular body}'' of unseen class \texttt{horse} also appear in seen classes \texttt{dear} and \texttt{zebra}, respectively.
There are three core issues in document-based ZSL: 
(1) \textbf{Noisy documents}: Unlike annotated attributes containing discriminative properties, encyclopedia documents are instead lots of noisy information that is irrelevant to recognizing the objects, such as descriptions about sound, diet, and organ structure (see red words in Figure \ref{fig: motivation}(b)). After recognizing noise suppression is crucial in this scenario, previous methods usually incorporate complex designs of regularizers (\textit{e.g.}, $L_{1,2}$ norm) ~\cite{ICML_2015_Bernardino, Document-based_CVPR_2016_Ruizhi_Qiao, Document-based_CVPR_2017_Mohamed} or pass textual features through additional fully connected (FC) layer~\cite{GAZSL, CIZSL, LsrGAN, I2DFormer, I2MVFormer} in model training. However, noisy information remains in the document, which harms knowledge transfer.
(2) \textbf{Less-described category documents}: Some Wikipedia articles, especially about fine-grained categories, are somewhat visually relevant but very short, \textit{e.g.}, dogs \texttt{chihuahua} and \texttt{collie} in AWA2~\cite{AWA2}, most classes in CUB~\cite{CUB_and_attribute_provide} and FLO~\cite{FLO} datasets. These documents lack sufficient semantic information to align images and text in the seen domain and recognize unseen classes. Although Reed et al.~\cite{document_as_AUX_1} use the Amazon Mechanical Turk (AMT) platform to collect detailed visual descriptions for each image in CUB and FLO, it is still labor-intensive. Recent work~\cite{I2MVFormer} leverages large language models (LLMs) to rewrite articles in different styles. However, this approach ignores that rewritten documents are still less described. How to enrich fine-grained category documents with minimal labor is challenging.
(3) \textbf{Suboptimal semantic alignment}: It is crucial for knowledge transfer to align visual words with corresponding image regions in document-based ZSL. Recent methods \cite{I2DFormer, I2MVFormer} enhance the semantic alignment by fine-grained interactions between image patches and words (or documents). However, they align noisy documents with image regions yet solely depend on attention mechanisms to implicitly select discriminative information. This process may fail to filter noisy information reliably and incorrectly relate image regions to non-visual words, as shown in Figure~\ref{fig: motivation}(b). Besides, the information is coupled in documents since they contain rich semantics, such as shape, color, and habitat. It is not easy to associate coupled semantics with corresponding regions accurately. The above issues harm knowledge transfer and limit the performance of previous methods.

In this work, we propose a \textbf{M}ulti-\textbf{A}ttribute \textbf{D}ocument \textbf{S}upervision (MADS) framework to address issues caused by noisy information at both document collection and model training stages.
As shown in Figure \ref{fig: motivation}(c), inspired by recent advancements in large language models (LLMs)~\cite{ChatGPT, GPT4_Tech}, we exploit its extensive world knowledge by a novel prompt algorithm to automatically remove most noisy descriptions and decouple semantic information in documents (in Section~\ref{document_collection}).
Instead of directly querying LLMs to filter out noises, we convert this complex task into several simpler sub-tasks to improve noise suppression accuracy.
Specifically, we ask LLMs to act as domain experts to provide potential visual attribute views useful for recognizing categories in downstream tasks, \textit{e.g.}, \texttt{Shape, Color, and Pattern}.
This simple meta-process effectively helps LLMs reduce the number of visual descriptions mistaken for noises.
Then, collected documents are divided into paragraphs based on these views, which filters out noisy sentences that do not describe the above views.
We also enrich the descriptions of the less-described view with LLMs to obtain rich semantics for each class.
Subsequently, we introduce a novel MADS network to extract transferable knowledge from multi-attribute documents and achieve semantic alignment between the visual words of each view and corresponding image regions (in Section~\ref{model_architecture}).
Specifically, we first represent the core semantics of each attribute paragraph to obtain view embeddings by the independent view semantic extraction module.
Unlike implicit attention mechanisms~\cite{I2DFormer, I2MVFormer}, a focus loss is proposed to penalize view embeddings, explicitly decreasing attention on non-visual words. 
Since semantic association appears between distinct views, such as ``\texttt{black marks}'' in color is details of ``\texttt{muscular body}'' in shape, we propose a multi-view semantic aggregation module to enhance interactions among view embeddings. 
After that, we introduce two losses to align view embeddings of each attribute paragraph with corresponding visual regions at global and local levels, which also provides interpretable predictions from multiple attribute views.

Our key contributions are as follows.
(1) To address issues caused by noisy information in document-based ZSL, we propose a \textbf{M}ulti-\textbf{A}ttribute \textbf{D}ocument \textbf{S}upervision (MADS) framework to remove noise in both document collection and model training. This ensures the model is trained with cleaner auxiliary information approximating attributes.
(2) We design a novel prompt algorithm to instruct LLMs to select visually descriptive sentences from multiple attribute views into paragraphs to decouple semantics before model training. To our knowledge, it is the first algorithm to reduce noisy descriptions of documents without human-in-the-loop.
(3) We introduce the MADS network to extract discriminative and transferable knowledge from multi-attribute documents and achieve semantic alignment between visual words in each attribute view and corresponding image regions. A model-agnostic focus loss is proposed to encourage the model to explicitly attend to visual words, which also improves previous methods without additional parameters.
(4) With comparable computation costs, our model consistently outperforms the SOTA by 7.2\% and 8.2\% across three benchmarks for document-based ZSL and GZSL settings, respectively. 
Moreover, we qualitatively explain the performance gains by accurate semantic alignment and interpretable predictions from different views.

\section{Related Work}
\label{related_work}
\subsection{Zero-Shot Learning}
Zero-Shot Learning (ZSL) aims to train on seen classes and generalize to recognize unseen classes~\cite{ZSL_invent1, ZSL_invent2, ZSL_invent3} during the test stage. 
Using auxiliary information to transfer knowledge is the key to the ZSL.
Most methods rely on annotated attributes ~\cite{AWA_attribute_provide2, AWA1_and_attribute_provide, CUB_and_attribute_provide} as auxiliary information.
Early embedding-based methods~\cite{CVPR_2013_Zeynep, TPAMI_2015_Zeynep_Akata, ICML_2015_Bernardino, CVPR_2016_Soravit_Changpinyo, CVPR_2016_Yongqin_Xian} map visual and auxiliary embeddings into a common semantic space by compatibility functions.
Recent methods build fine-grained interactions between attributes and image regions~\cite{NIPS_2018_Shichen_Liu, ICCV_2019_Huajie_Jiang, CVPR_2021_Zongyan_Han, APN, CVPR_2022_MSDN} or extract attribute-guided visual features with attention mechanisms and strong visual backbones~\cite{NIPS_2018_Yunlong_Yu, CVPR_2021_Yang_Liu, AAAI_2023_Duet, semantic_align_TransZero, CVPR_2023_PSVMA, semantic_align_ZSLViT, Embed_TIP_2024} to enhance semantic alignment.
Generative methods\cite{CVPR_2018_Yongqin_Xian, CVPR_2019_f_vaegan_d2, ICCV_2019_Yizhe_zhu, CVPR_2022_En-Compactness, CVPR_2022_Hongzu_Su, GEN_method_VS_Boost, GZSL_ICML_2023, CVPR_2024_GEN_1, GEN_TIP_2022, GEN_TIP_2024} covert the ZSL task to supervised learning by training a generator conditioned by semantic embeddings and synthesizing visual features for unseen classes. 
However, annotating attributes needs enormous human resources and deep domain expertise~\cite{Attribute_drawback_CVPR_2013, Attribute_drawback_ECCV_2018}. Moreover, more attributes are required to discriminate as novel classes increase~\cite{Attribute_drawback_and_document_arxiv_2021}. 
In contrast, other approaches ~\cite{word_embed_NIPS_2013_Richard_Socher, word_embed_NIPS_2013_Devise, SJE, word_embed_and_graph_CVPR_2019_Michael_Kampffmeyer, word_embed_and_graph_HGR, VGSE} utilize word embeddings from language models \cite{Glove, Word_embed_as_AUX_1, pre_trained_LM_EMNLP_2020_Wikipedia2vec} as auxiliary information, which are easily available.
Existing methods enhance semantic connections between class embeddings through knowledge graphs and hierarchical classifications\cite{word_embed_and_graph_CVPR_2018_Xiaolong_Wang, word_embed_and_graph_CVPR_2019_Michael_Kampffmeyer, word_embed_and_graph_CVPR_2021_Muhammad_Ferjad_Naeem, word_embed_and_graph_HGR}. 
However, since word embeddings are sensitive to linguistic issues and little discriminative information~\cite{Attribute_drawback_and_document_arxiv_2021, I2DFormer}, especially for fine-grained recognition, these methods achieve poor performance. Instead, textual documents about categories are easy to collect from many existing encyclopedias (\textit{e.g.}, Wikipedia). They contain rich information from domain knowledge, also with noisy and non-visual descriptions. In this work, we improve knowledge transfer for document-based ZSL by automatically filtering out noises in document collection and model learning. 


\subsection{Document-based Zero-Shot Learning}
Document-based ZSL aims to use category-level documents as auxiliary information, which transfers knowledge by shared visual-relevant descriptions between seen and unseen classes. Unlike annotated attributes, there are noisy descriptions and non-visual words in documents~\cite{Document-based_NAACL_2021_Jihyung_Kil, Attribute_drawback_and_document_arxiv_2021}. Recent work makes efforts to remove noisy information and achieve semantic alignment between visual and textual spaces. For noise suppression, early work~\cite{Document-based_ICCV_2013_Mohamed_Elhoseiny, ICML_2015_Bernardino, Document-based_CVPR_2016_Ruizhi_Qiao, Document-based_CVPR_2017_Mohamed} typically uses TF-IDF features~\cite{TF-IDF} as textual representations to filter out common but less informative words. They usually incorporate complex designs of regularizers (\textit{e.g.}, $L_{1,2}$ norm) ~\cite{ICML_2015_Bernardino, Document-based_CVPR_2016_Ruizhi_Qiao, Document-based_CVPR_2017_Mohamed} to encourage group sparsity on the connections to the textual terms. Recent work finds that an additional fully connected (FC) layer~\cite{GAZSL, CIZSL, LsrGAN} or attention mechanisms~\cite{I2DFormer, I2MVFormer} can dynamically weight important features and suppress noise.
However, noisy descriptions still exist, which harms knowledge transfer by introducing irrelevant or misleading features that obscure the most discriminative information. In this work, we propose multi-attribute document supervision to remove noisy descriptions and decouple information without human intervention before model training. 

To achieve semantic alignment, most work~\cite{Document-based_ICCV_2015_Lei_Jimmy_Ba, Document-based_CVPR_2016_Ruizhi_Qiao, Document-based_CVPR_2017_Ziad} directly relates document embeddings by TF-IDF ~\cite{TF-IDF} or language models~\cite{word2vec, document_as_AUX_1, LSTM, BERT} with image features. Several methods ~\cite{Document-based_CVPR_2017_Mohamed, GAZSL, CIZSL, LsrGAN} extract salient regions by part detection network to enhance image embeddings. However, annotated attributes are used to train the detection model.
Recently, some work ~\cite{I2DFormer, I2MVFormer} learns fine-grained interactions to enhance semantic alignment. Specifically, I2DFromer ~\cite{I2DFormer} relates image patches with words in global and local compatibility. I2MVFormer ~\cite{I2MVFormer} aggregates information at the document level to reduce computation cost. However, these methods aggregate coupled semantics and filter noisy information yet solely depend on the implicit attention mechanism, which may incorrectly relate image regions to non-visual words. 
In this work, we associate the decoupled semantic embedding of each view with image local features to enhance fine-grained alignment. Besides, a model-agnostic focus loss is proposed to attend to visual words explicitly during the training.

\subsection{Image Classification with Descriptions Generated by LLMs}
Recent work shows that image classification with descriptions generated by LLMs can improve the model generalization ability. Several methods ~\cite{DCLIP, CuPL, GPT4Vis, VDT_2023_ICCV_gpt4, MPVR} enhance large vision-language models (VLMs) like CLIP~\cite{CLIP} by using visual descriptions of each class to language prompts, \textit{e.g.}, ``\texttt{a photo of a \{classname\}. \{description\}.}''. DCLIP ~\cite{DCLIP} and CuPL ~\cite{CuPL} first instruct LLMs to generate category descriptions by query prompts such as ``\texttt{What does a category look like?}''. Recent methods ~\cite{GPT4Vis, VDT_2023_ICCV_gpt4} utilize similar query prompts to ask stronger LLMs, \textit{i.e.}, GPT-4 ~\cite{GPT4_Tech}, for rich and diverse descriptions. MPVR ~\cite{MPVR} designs a meta-prompt to produce task-specific query prompts. However, these methods rely heavily on the prior information of class names in VLMs (see Section~\ref{sec:compare_with_clip}).
In contrast, I2MVFormer ~\cite{I2MVFormer} asks LLMs to provide additional auxiliary information in specific domains. It rewrites category documents in various styles to enrich semantic knowledge. However, this method ignores that documents still contain non-visual and less-described sentences, which rewriting cannot alleviate. 
In this work, we aim to improve the quality of auxiliary information, \textit{i.e.}, category documents. A novel prompt algorithm is proposed to remove noisy information and enrich less-described categories in encyclopedia articles. 
This process ensures that the models are trained under conditions that approximate the attributes, \textit{i.e.}, supervision of human experts.

\subsection{Fine-grained Semantic Localization}
In attribute-based ZSL, some works~\cite{APN, DAZLE, semantic_align_TransZero, RGAT} show that attribute localization is helpful for fine-grained semantic alignment and knowledge transfer.
APN~\cite{APN} uses an attribute prototype network to regress and decorate attributes from intermediate features.
DAZLE~\cite{DAZLE} applies a dense attribute-based attention mechanism to focus on the most relevant image regions. 
Transzero~\cite{semantic_align_TransZero} proposes an attribute-guided transformer to reﬁne visual features and learn attribute localization for discriminative representations. 
RGAT~\cite{RGAT} jointly learns attribute-region embeddings and incorporates global embeddings under the guidance of a graph to capture relationships between regions. 
However, unlike annotated attributes, documents have noisy descriptions and non-visual words. We introduce multi-attribute document supervision to remove noise and focus loss to attend to visual words rather than all words. In contrast, attribute localization work does not necessitate considering this challenge, directly relating regions with all annotated attributes. Besides, we align local and global image features with semantic embeddings from multiple decoupled views to alleviate the bias to seen classes.

{
\begin{figure*}[tb]
\begin{center}
\centerline{\includegraphics[width=0.85\linewidth]{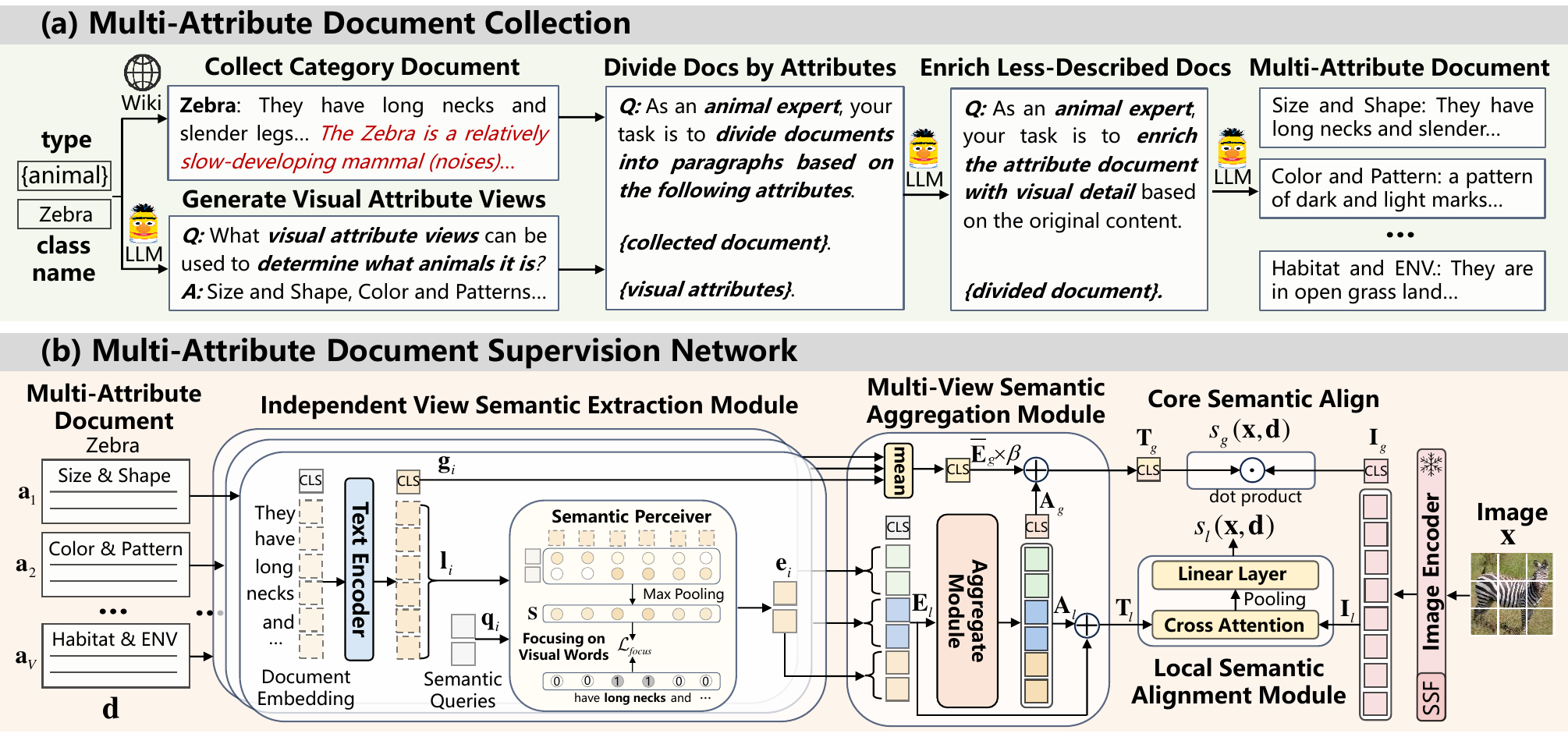}}
\vskip -0.09in
\caption{ An overview of our framework.
(a) Document Collection. 
We instruct LLMs to divide the definition document into paragraphs based on attribute views and enrich the less-described attribute documents.
(b) Our MADS Network. 
The MADS first extracts the core semantics of each attribute paragraph independently and aggregates multi-view semantics to enhance interactions, aligning with global and local image embeddings. 
Unlike previous methods that solely depend on implicit attention mechanisms, we introduce a focus loss to explicitly filter noisy information and attend to visual words.
}
\label{fig: Method}
\end{center}
\vskip -0.16in
\end{figure*}
}


\section{Multi-Attribute Document Collection}
\label{document_collection}


In document-based ZSL, category documents are the theoretical foundation for knowledge transfer. 
Compared to previous methods \cite{I2DFormer, I2MVFormer}, we collect multi-attribute documents with information decoupling as model input instead of noisy documents. 
Inspired by decomposed prompting ~\cite{decomposed_prompt, decomposed_prompt_EMNLP} in natural language processing, we convert the complex task of multi-attribute document collection to simpler sub-tasks that can easily be delegated to a library of prompting-based LLMs.
As shown in Figure~\ref{fig: Method}(a), we first collect category documents from encyclopedias (\textit{e.g., Wikipedia}) and ask LLMs to generate potential visual attribute views that can be used to recognize specific species. 
Leveraging robust text understanding abilities by LLMs, we divide the documents based on attribute views and further enrich the less-described attribute documents.
We illustrate detailed processes in the Algorithm~\ref{alg: collect_document}, which will be explained in the following paragraphs.

\textbf{Collecting Category Documents.}
Similar to \cite{I2DFormer, I2MVFormer}, we collect documents from encyclopedias, \textit{i.e.}, the A-Z animals \cite{azanimal}, All About Birds~\cite{aab} and Wikipedia~\cite{wiki} for AWA2~\cite{AWA2}, CUB ~\cite{CUB_and_attribute_provide} and FLO~\cite{FLO} datasets, respectively. 
However, since documents in encyclopedias are used to define a category, they contain exhaustive descriptions with visual (\textit{e.g.}, \texttt{appearance}, \texttt{habitat}, and \texttt{behavior}) and non-visual (\textit{e.g.}, \texttt{sound}, \texttt{diet}, and \texttt{organ structure}) information.
It is difficult to avoid introducing noise in this process. 
Moreover, although we search from multi-web sources, some documents about fine-grained classes still have few visual-relevant descriptions, \textit{e.g.}, dogs \texttt{chihuahua} and \texttt{collie} in AWA2~\cite{AWA2}, most classes in CUB~\cite{CUB_and_attribute_provide} and FLO~\cite{FLO}.
In the subsequent processes, we will address these issues.
Ultimately, we organize collected descriptions for each category into one paragraph, denoted as \textit{collected document} later.


\textbf{Generating Visual Attribute Views.}
The collected document describes the category from multiple semantic views. We wish to select visually relevant descriptions from them.
Instead of directly asking LLMs to do this, we introduce a meta-process to help LLMs attend to potential visual attribute views that can be used to describe the visual details, such as \texttt{Size and Shape}, \texttt{Color and Pattern}, and \texttt{Habitat and Environment}.
We empirically find that this process reduces the number of visual descriptions mistaken for noises. Meanwhile, these attributes provide a basis for semantic decoupling in documents.
For this purpose, we use the prompt $p_{view}$:
\begin{quote}
    \makebox[\linewidth]{%
        \colorbox{lightblue}{%
            \hspace*{0mm} 
            \begin{minipage}{\dimexpr\linewidth+10\fboxsep\relax} 
                \fontsize{9pt}{10pt}\selectfont 
                ``Given an image, what visual attribute views can be used to determine what \{\texttt{type}\} species it is?''
            \end{minipage}%
        }%
    }
\end{quote}
\noindent Here, \{\texttt{type}\} is the dataset domain, which provides task-specific descriptions to guide LLMs into expert roles. We use \textit{``animal''}, \textit{``bird''} and \textit{``flower''} as \{\texttt{type}\} for AWA2, CUB, and FLO datasets, respectively.
However, due to output diversity in LLMs, we obtain different responses at each query time. To this end, we query five times (\textit{i.e.}, $T_{repeat} = 5$) and select the common visual attributes as the final result.


\textbf{Dividing Documents by Visual Attributes.}
After obtaining a set of visual attributes, we instruct LLMs to act as domain experts, dividing the definition document into paragraphs based on these attributes. We query an LLM as prompt $p_{divide}$:
\begin{quote}
    \makebox[\linewidth]{%
        \colorbox{lightblue}{%
            \hspace*{0mm} 
            \begin{minipage}{\dimexpr\linewidth+10\fboxsep\relax} 
                \fontsize{9pt}{10pt}\selectfont 
                ``As an \{\texttt{type}\} expert, your task is to divide documents into paragraphs based on the following attribute views. Each paragraph contains descriptions of one attribute view. \{\texttt{collected document}\}. \{\texttt{visual attributes}\}.''
            \end{minipage}%
        }%
    }
\end{quote}
\noindent Here, \{\texttt{type}\} is also the dataset domain. \{\texttt{collected document}\} and \{\texttt{visual attributes}\} are obtained from the above two processes.
This process outputs original attribute documents with multiple paragraphs, denoted as \textit{divided document} later.
Since these views are visually related, this process effectively filters out non-visual descriptions. Meanwhile, each paragraph only contains visual sentences of the current attribute view, which achieves information decoupling and is helpful for accurate semantic alignment.

\textbf{Enriching Less-Described Attribute Documents.}
While encyclopedias offer comprehensive definitions for categories, collected documents sometimes lack descriptions for some attributes.
Inspired by extensive human knowledge (trained on web-scale text data) in LLMs, we enrich less-described attribute documents with the following prompt $p_{enrich}$:
\begin{quote}
    \makebox[\linewidth]{%
        \colorbox{lightblue}{%
            \hspace*{0mm} 
            \begin{minipage}{\dimexpr\linewidth+10\fboxsep\relax} 
                \fontsize{9pt}{10pt}\selectfont 
                ``As an \{\texttt{type}\} expert, your task is to enrich the attribute documents with concise visual details.
                Try to keep the original description unchanged. 
                Below is the document about \{\texttt{category}\} that needs additional descriptions: \{\texttt{divided document}\}.''
            \end{minipage}%
        }%
    }
\end{quote}
Here, \{\texttt{category}\} and \{\texttt{divided document}\} are the class names in datasets and corresponding divided documents. 
\{\texttt{type}\} is the dataset domain, which provides task-specific context information to reduce semantic ambiguity in category names.
We query LLMs to generate concise visual details for missing attributes instead of any details, avoiding the non-visual descriptions generated by LLMs. 
Moreover, we ask LLMs to keep the documents unchanged, which ensures that the original visual knowledge in encyclopedia is preserved. 
We empirically find that enriching descriptions with divided documents as inputs result in more detailed and longer descriptions than directly to collected documents.

\begin{algorithm}[t]
\caption{Multi-Attribute Document Collection Algorithm}
\label{alg: collect_document}
\begin{algorithmic}[1]
\REQUIRE repeated times $T_{repeat}$, class labels $\mathcal{C}$, LLM \texttt{LLM}, dataset domain $\mathcal{T}$, Encyclopedia \texttt{Pedia}
\STATE Initialize visual attribute set: $\mathcal{A} \leftarrow \{\}$
\FOR{$t=1$ to $T_{repeat}$}
    \STATE Generate visual attribute views: $\mathcal{A}_t \leftarrow$ \texttt{LLM} ($p_{view}, \mathcal{T}$)
    \STATE Update visual attribute set: $\mathcal{A} \leftarrow \mathcal{A} \cap \mathcal{A}_t$ 
\ENDFOR
\STATE Initialize document set: $\mathcal{M} \leftarrow \{\}$
\FORALL{class $c_i \in \mathcal{C}$}
    \STATE Collect document for category $c_i$: $D_{i} \leftarrow \texttt{Pedia} (c_i)$
    \STATE Divide documents by visual attribute views: $D_{i} \leftarrow$  \texttt{LLM} ($p_{divide}, \mathcal{T}, D_i, \mathcal{A} $)
    \STATE Enrich document for less-described attributes: $D_i \leftarrow$ \texttt{LLM} ($p_{enrich}, \mathcal{T}, D_i, c_i$)
    \STATE Update document set: $\mathcal{M} \leftarrow \mathcal{M} \cup  D_{i}$
\ENDFOR
\RETURN multi-attribute document set $\mathcal{M}$
\end{algorithmic}
\end{algorithm}

After the above processes, we collect detailed visual descriptions of multiple attribute views for each category, denoted as \textit{multi-attribute document} later. 
These documents serve as auxiliary information for subsequent model training. 
Notably, LLMs are solely involved in document collection and are not used in model training.
In Section~\ref{sec: qualitative_analysis}, we qualitatively visualize this process and demonstrate its effectiveness.

\section{Multi-Attribute Document Supervision}
\label{model_architecture}

We propose the \textbf{M}ulti-\textbf{A}ttribute \textbf{D}ocument \textbf{S}upervision (MADS) network in Figure~\ref{fig: Method}(b).
The MADS first learns multiple semantic embeddings by an independent view semantic extraction module and a multi-view semantic aggregation module (see Section~\ref{sec: obtain_semantic}), which extracts discriminative and transferable information from documents.
To explicitly filter noisy information, we introduce a focus loss to penalize the model for paying more attention to task-helpful visual words rather than all words when computing the attention mechanism (see Section~\ref{sec: focus_loss}).
Then, multiple semantic embeddings are aligned to salient image features at global and local levels by core and local semantic alignment losses (see Section~\ref{sec: semantic_alignment}).


\textbf{Notation.} 
ZSL aims to recognize unseen classes $\mathcal{Y}^u$ without any labeled data by training a model solely on images from seen classes $\mathcal{Y}^s$.
Here, $\mathcal{Y}^s \cap \mathcal{Y}^u = \emptyset$, \textit{i.e.}, the unseen classes are disjoint from seen classes.
In this work, we use multi-attribute documents as auxiliary information to transfer knowledge.
Each class (seen or unseen) has a corresponding document $\bm d = \{{{\bm a}_1, {\bm a}_2, \cdots, {\bm a}_V} \}$ that describes it from multiple attribute views, where $V$ is the number of views.
The training set is defined as $ \mathcal{S} = \{ (\bm x, \bm y, \bm d) | \bm x \in \mathcal{X}^s, \bm y \in \mathcal{Y}^s,  \bm d \in \mathcal{D}^s \}$, where $\bm x$ is an image from the training data $\mathcal{X}^s$, $\bm y$ is the corresponding label from seen classes  $\mathcal{Y}^s$, $\bm d$ is the associated document.
Notably, all training data belongs to seen classes.
At test time, we evaluate the model on another set of images $\mathcal{X}_{test}$ and their corresponding classes $\mathcal{Y}_{test}$ and documents $\mathcal{D}_{test}$.
For the ZSL setting, test images are only from unseen classes (\textit{i.e.}, $\mathcal{Y}_{test} = \mathcal{Y}^{u}$). For generalized ZSL, test images are from both seen and unseen classes (\textit{i.e.}, $\mathcal{Y}_{test} = \mathcal{Y}^s \cup \mathcal{Y}^u)$.


\subsection{Learning Multi-View Semantic Embeddings}
\label{sec: obtain_semantic}
With multi-attribute documents as input, the new challenge is how to extract visual-semantic knowledge from them. We first consider whether the SOTA models address this issue, \textit{i.e.}, I2DFormer~\cite{I2DFormer} and I2MVFormer~\cite{I2MVFormer}. 
As I2DFormer~\cite{I2DFormer} uses a single document as input, we can integrate multi-attribute documents into one paragraph and extract semantic embeddings. 
However, this approach associates coupled semantics with image regions, which makes it hard to achieve accurate semantic alignment.
In contrast, I2MVFormer \cite{I2MVFormer} supports multiple paragraphs as input, encoding each paragraph to a set of view embeddings and aggregating them by a mean operation.
However, I2MVFormer ignores the varied contributions of each attribute paragraph for image recognition, treating each view embedding equally.
Moreover, semantic associations exist among different attribute views.
For example, \texttt{dark marks} in color and \texttt{coarse fur} in texture are detailed descriptions of \texttt{muscular body} in shape.
This approach also fails to capture the semantic interaction.

To address these issues, we first introduce the independent view semantic extraction module to represent the core semantics of each view in documents. This process encodes each paragraph independently to ensure information decoupling.
Then, we propose the multi-view semantic aggregation module to enhance the semantic interaction between different attribute views at global and local levels.
The details are as follows.


\textbf{Independent View Semantic Extraction.}
Given $\bm d = \{{{\bm a}_1, {\bm a}_2, \cdots, {\bm a}_V} \}$, the multi-attribute document of class $\bm y$,  we initialize words with Glove embedding~\cite{Glove} similar to~\cite{I2DFormer, I2MVFormer}.
Then, we pass them to a learnable shallow MLP layer to improve word embeddings and reduce the feature dimension to $r$. For each attribute paragraph ${\bm a}_i$, the output of this yields ${\bm w}_i \in \mathbb{R}^{M \times r}$, where $M$ is the length of the paragraph:
\vspace{-2pt}
\begin{equation}
    {\bm w}_i =  {\bm W}_2 \cdot \delta ({\bm W}_1 \cdot \text{Glove}({\bm a}_i)), \; \text{where}\; i = 1, 2, \cdots, V.
\end{equation}
\noindent
Here, ${\bm W}_1$ and ${\bm W}_2$ are learnable matrices of weights in MLP, and $\delta(\cdot)$ is the activation function. 
We append a [\textit{CLS}] token ${\bm c}_i \in \mathbb{R}^r$ to each sequence ${\bm w}_i$ and feed them to a text encoder $\phi(\cdot)$, which contains transformer encoder blocks training from scratch. 
For a single sequence ${\bm w}_i$, the text encoder outputs $ \bm{g}_i \in \mathbb{R}^{r} $ in [\textit{CLS}] position as the single-view core feature and $ \bm{l}_i \in \mathbb{R}^{M \times r}$ in other positions as local text features:
\begin{equation}
    [\bm{g}_i, \bm{l}_i] =  \phi([{\bm c}_i, {\bm w}_i]).
\end{equation}

The average size of an attribute paragraph is about 150 words, which is more challenging than CLIP~\cite{CLIP} with a max of 76 tokens.
To reduce memory costs, we apply a semantic perceiver to extract salient information in local text features by the cross-attention mechanism.
For each attribute view, the perceiver represents local text features ${\bm l}_i$ as key and value and a set of learnable tokens ${\bm q}_i \in \mathbb{R}^{K \times r}$ as semantic queries, where $K$ is much smaller than the number of words $M$.
Three learnable projection layers ${\bm W}_k^c, {\bm W}_q^c$ and ${\bm W}_v^c$ are introduced to map key, query, and value to new spaces with dimension $r_h$.
Then, we use the dot-product attention mechanism to aggregate information in ${\bm l}_i$, formally defined as:
\begin{align}
    {\bm K}_i = {\bm W}_k^c {\bm l}_i, &\; {\bm Q}_i = {\bm W}_q^c {\bm q}_i, \; {\bm V}_i = {\bm W}_v^c {\bm l}_i,  \\
    {\bm H}_i  &= \text{softmax} \bigg (  \frac{ \bm{Q}_i  \bm{K}_{i}^T }{\sqrt{r_h}} \bigg ), \\
    {\bm e}_i &= {\bm H}_i {\bm V}_i  {\bm W}_o^c,
\end{align}
where ${\bm W}_o^c \in \mathbb{R}^{r_h \times r}$ is a learnable linear layer to map the feature to the original dimension. This process outputs ${\bm e}_i \in  \mathbb{R}^{K \times r} $ as single-view salient local features.
Finally, we obtain the single-view core feature ${\bm g}_i$ and salient local features ${\bm e}_i$ for each attribute paragraph, where $i = 1, 2, \cdots, V$.


\textbf{Multi-View Semantic Aggregation.}
To fuse the semantic information and consider the varied recognition contributions between different views, a multi-view semantic aggregation module is introduced to build interactions at global and local levels. 
We first concatenate the salient local features as ${\bm E}_{l} \in \mathbb{R}^{(V \cdot K) \times r}$ and append a learnable [\textit{CLS}] token ${\bm z}$:
\begin{equation}
    {\bm E}_{l} = [{\bm e}_1, {\bm e}_2, \cdots, {\bm e}_V], \; \hat {\bm E} = [\bm z, {\bm E}_{l}].
\end{equation}
We input this sequence $\hat {\bm E}$ to an aggregate module containing a self-attention layer and a shallow MLP layer. The aggregate module outputs $\bm{A}_g \in \mathbb{R}^{r} $ at [\textit{CLS}] token as the aggregated global feature and $\bm{A}_l \in \mathbb{R}^{(V \cdot K) \times r}$ at other words tokens as aggregate local features, formulated as:
\begin{align}
    {\bm K}_a = {\bm W}_k^s  \hat {\bm E}, \; {\bm Q}_a &= {\bm W}_q^s  \hat {\bm E}, \; {\bm V}_a = {\bm W}_v^s  \hat {\bm E}, \\
    \tilde {\bm E} = \text{softmax} &\bigg(\frac{{\bm Q}_a {\bm K}_a^T}{\sqrt{r_h}} \bigg ) {\bm V}_a {\bm W}_o^s, \\
    [\bm{A}_g, \bm{A}_l]  &=  \tilde {\bm E} + \text{MLP}(\tilde {\bm E}).
\end{align}
Here, ${\bm W}^s_k$, ${\bm W}^s_q$, ${\bm W}^s_v$ are three projection layers to map the sequence to different spaces. ${\bm W}_o^s \in \mathbb{R}^{r_h \times r}$ maps the feature to the original dimension. 
To maintain decoupled semantics of each view in the global feature, we pass all single-view core features with a mean operation, denoted as $\overline{\bm{E}}_g \in \mathbb{R}^{r}$, and fuse it with the aggregated global feature by a weighted sum:
\vspace{-4pt}
\begin{align}
    \overline{\bm{E}}_g &= \frac{1}{V} \sum_{i=1}^V {\bm g}_i, \\
    \bm{T}_{g} &= \beta \cdot \overline{\bm{E}}_g  + (1 - \beta) \cdot \bm{A}_g, 
    \label{eq:weighed_sum}  
\end{align}
where the scalar $\beta$ is the fusion ratio to balance independent and aggregated features. $\beta = 1$ means treating each attribute view equally without considering semantic interactions.
Similarly, we also fuse aggregated with single-view local features by a residual connect to maintain information decoupling,
\vspace{-2pt}
\begin{equation}
    \bm{T}_{l} = \bm{E}_{l} + \bm{A}_l.
    \label{eq:res_connect}
\end{equation}
We denote ${\bm T}_g$ and ${\bm T}_l$ as global and local semantic embeddings.

\subsection{Focusing Attention on Visual Words}
\label{sec: focus_loss}
While we filtered noisy sentences, words without visual information, such as prepositions and adverbs, may still remain. 
As shown in Figure \ref{fig: motivation}(a), visual words (\textit{i.e.}, adjectives and nouns that describe visual details) are the bridge to transfer knowledge.
To this end, we introduce a focus loss to help the semantic perceiver enhance attention to visual words.

A challenge lies in automatically recognizing whose visual words are in documents.
We argue that labels in recent object recognition tasks, such as Recognize Anything Model (RAM)~\cite{RAM_label} and Visual Genome (VG)~\cite{VG_dataset}, consist of detailed visual nouns and adjectives covering various aspects of animals, plants, and birds, etc. Moreover, these labels describe color, shape, habitat, and object parts, which contain the most common attribute views in our documents.
Therefore, we define words appearing in these labels as visual words:
\begin{equation}
    \psi(\bm t) = \begin{cases}
    1 & \text{if} \;t\;\text{in RAM or VG labels}, \\
    0 & \text{otherwise.}
    \end{cases}
\end{equation}
Here, ${\bm t}$ is a word, and $\psi(\bm t) = 1$ denotes it is a visual word.


To enhance attention to visual words, an easy idea is to retain them and remove other words in documents.
However, we empirically find that this naive idea significantly degrades performance (see Section~\ref{sec: ablate_focus_loss}).
This may be because other words provide contextual information for visual words, assigning different values like attribute values.
Therefore, we keep other words in documents and penalize the semantic perceiver to focus on visual words in the attention operation.
The maximum attention score of each word $\mathcal{S} = \{ {\bm s}_1, {\bm s}_2, \cdots, {\bm s}_M \}$ is obtained by a max pooling on the attention map ${\bm H} \in \mathbb{R}^{K \times M}$, where $K$ and $M$ is the number of semantic queries and word tokens, respectively.
We penalize the maximum attention score of visual words to be 1 and otherwise 0 by the following loss:
\begin{align}
    \label{eq: focus_loss}
    \varphi({\bm H}) = \sum_{j=1}^M [\psi ({\bm t}_j) \cdot \text{log}&({\bm s}_j) + (1 - \psi ({\bm t}_j)) \cdot \text{log}({\bm s}_j)], \\
    \mathcal{L}_{focus} &= \frac{1}{V} \sum_{i=1}^V \varphi({\bm H}_i),
\end{align}
where ${\bm t}_j$ is the word in a attribute paragraph, and ${\bm s}_j$ is corresponding maximum attention score.
Since the loss only penalizes the attention mechanism without additional parameters, \textit{i.e.}, model-agnostic, our analysis (see Section~\ref{exp: improve_previous_methods}) demonstrates that it is also helpful to previous methods.





\subsection{Aligning Multiple Semantic Embeddings with Images}
\label{sec: semantic_alignment}
To better transfer knowledge, two losses are proposed to align semantic embeddings from multiple views with visual features at global and local levels. 
The core semantic alignment loss is proposed to align single-view core semantics and aggregated semantics with the global image feature.
The local semantic alignment loss aims to relate local semantic embeddings with discriminative region features, which offer helpful information for fine-grained classification.

\textbf{Image Encoder.} We leverage an image encoder with a few learnable parameters to select the salient visual information helpful for category recognition.
Given an input image $\bm x$, the image encoder extracts features by a fixed ViT \cite{vit}. To obtain better visual perception, we add shift and scale \cite{SSF} for each block output in ViT. 
A projection layer with dimension $r$ is applied to reduce feature dimension after the ViT. 
The image encoder outputs the $\bm{I}_g \in \mathbb{R}^r  $ at [\textit{CLS}] token as the global image feature and $\bm{I}_l \in \mathbb{R}^{N \times r}$ at other positions as patch-wise local image features, where $N$ is the number of image patches.

\textbf{Core Semantic Alignment Loss} penalizes global image feature $\bm{I}_g$ and corresponding global semantic embedding $\bm{T}_{g}$ to be closer than other pairs by the cross-entropy loss:
\begin{equation}
    \mathcal{L}_{global} = -\log \frac{\text{exp} (s_{g}(\bm x, \bm d) )}{\sum_{{\bm d}' \in \mathcal{D}^s} \text{exp} (s_{g}(\bm x, {\bm d}'))},
\end{equation}
where $s_{g} (\bm x, \bm d)$ is a global score, formulated as $s_{g} (\bm x, \bm d) = \bm{I}_g \cdot \bm{T}_{g}$.
Notably, since ${\bm T}_g$ in Eq.~\ref{eq:weighed_sum} contains core semantics for each attribute view as well as aggregated semantics with interactions, this process provides interpretable predictions by computing cosine similarity between visual-semantic pairs.

\textbf{Local Semantic Alignment Loss}.
Similar to~\cite{I2DFormer, I2MVFormer}, we introduce a module to encourage fine-grained semantic associations at the word-to-region level. 
This module applies a cross-attention block to enhance visual features with semantic knowledge in the document.
We feed local image features $\bm{I}_l$ as query and multi-view local semantic embeddings $\bm{T}_{l} $ as key and value to the block, yielding $\hat{\bm{I}}_l \in \mathbb{R} ^ {N \times r} $. 
The enhanced features $\hat{\bm{I}}_l$ are fed into a global pooling operation on patch dimension $N$ to aggregate discriminative features, followed by a linear layer $L (\cdot) \in \mathbb{R}^{r \times 1}$ to output a local score $s_{l} (\bm x, \bm d)$.
We optimized this score by a cross-entropy loss: 
\begin{align}
    s_{l} (\bm x, \bm d) &= L( \text{Pool}_N (\hat{\bm{I}}_l) )\\
    \mathcal{L}_{local} = -&\log \frac{\text{exp} (s_{l}(\bm x, \bm d) )}{\sum_{{\bm d}' \in \mathcal{D}^s} \text{exp} (s_{l}(\bm x, {\bm d}'))}.
\end{align}
Here, this objective is to encourage salient regions in $\bm x$ to be close to local semantics in its corresponding document $\bm d$.
Since ${\bm T}_l$ in Eq.~\ref{eq:res_connect} extracts salient information in each attribute paragraph, especially about visual words, this process also achieves semantic alignment between the visual words of each view and corresponding image regions.

We optimize our MADS network with the following loss:
\begin{equation}
    \label{eq: final_loss}
    \mathcal{L} = \mathcal{L}_{global} + \lambda_{local} \mathcal{L}_{local} + \lambda_{focus} \mathcal{L}_{focus},
\end{equation}
where $\lambda_{local}$ and $\lambda_{focus}$ are hyper-parameters. 

\subsection{Inference}
\label{sec:train_and_inference}

Given an input image $\bm x$, we obtain a prediction $\hat {\bm y}$ that yields the highest global score among unseen classes for ZSL:
\begin{equation}
    \hat {\bm y} = \argmax_{{\bm d}' \in {\mathcal{D}^u}} \; s_{g}(\bm x, {\bm d}').
\end{equation}

For GZSL, we compute the global score among both seen and unseen classes.
Moreover, we also apply calibrated stacking (CS) \cite{calibrated_stack} strategy to trade-off calibration degrees, reducing the bias towards seen classes. It is formulated as:
\begin{equation}
    \hat {\bm y} = \argmax_{{\bm d}' \in {\mathcal{D}^s \cup \mathcal{D}^u}} \; ( s_{g}(\bm x, {\bm d}') - \gamma \mathbb{I}_{\mathcal D^s} ({\bm d}') ).
\end{equation}
Here, $\mathbb{I}_{\mathcal D^s}$ is an indicator function, which is 1 when  ${\boldsymbol d}' \in \mathcal{D}^s$ and 0 otherwise. A scalar factor $\gamma$ is to trade off the calibration degree on seen classes. Notably, as global features have fused with fine-grained knowledge, we only use the global score rather than the local score to save costs during inference.

\section{Experiments}
\label{experiment}


\textbf{Datasets.} 
We evaluate our model on three widely used datasets for ZSL, \textit{i.e.}, Animals with Attributes2 (AWA2)~\cite{AWA2}, Caltech-USCD Birds-200-2011 (CUB)~\cite{CUB_and_attribute_provide}, and Oxford Flowers (FLO)~\cite{FLO}. 
AWA2 is a coarse-grained animal recognition consisting of 37,322 images from 50 classes (seen/unseen classes = 40/10).
CUB is a fine-grained bird recognition, which has 11,788 images of 200 classes (seen/unseen classes = 150/50). 
FLO is a fine-grained flower recognition, containing 9,862 images from 102 classes (seen/unseen = 82/20).
We follow the Proposed Split (PS)~\cite{AWA2} for seen-unseen class division. 
Notably, we utilize our collected documents as auxiliary information instead of annotated attributes in datasets.

\textbf{Evaluation Metrics.} 
Following~\cite{AWA2}, we use the average per-class top-1 accuracy as the evaluation metric.
In the ZSL setting, test images are only from unseen classes.
We compute the accuracy of each unseen class and measure the average per-class top-1 accuracy (T1).
In the GZSL setting, the search space contains both seen and unseen classes.
We present the per-class mean accuracy on both seen ($S$) and unseen classes ($U$), as well as their harmonic mean $H = (2\times U \times S)/ (U + S) $.

\textbf{Implementation Details.}
We use the ViT-B/16~\cite{vit} pre-trained on ImageNet 1K~\cite{ImageNet} as the visual backbone.
This respects the GUB split~\cite{AWA2}, which means no unseen classes are used to train the visual backbone.
For a fair comparison, the dimension of semantic embedding $r$ is 256, 64, 128 for AWA2, CUB, and FLO, similar to~\cite{I2DFormer, I2MVFormer}.
We use two layers of text encoder, two layers of semantic perceiver, and one layer of aggregation module.
The model is trained with AdamW~\cite{Adamw} optimizer with a learning rate decreased cosine with each step. 
We optimize hyperparameters by grid search in the validation split. 
Once hyperparameters are confirmed, we merge the validation split with the training split.
We vary $\lambda_{focus} \in [0, 5.0]$, $\lambda_{local} \in [0, 1.0]$, fusion ratio $\beta \in [0, 1.0]$, the number of semantic queries $K \in [1, 10]$ and LLM temperature $\tau \in [0.6, 1.4]$ for training. 
The effect of hyperparameters is shown in Figure~\ref{fig:loss_influence}.
Besides, our MADS is implemented in PyTorch and trained with an Nvidia GeForce RTX 3090 GPU.

\subsection{Comparison With the State-of-the-Art Methods}

\textbf{Compared Methods}.
We compare our results with state-of-the-art (SOTA) ZSL methods with documents as auxiliary information.
We first compare the SOTA methods using word embeddings of class names, the simplest form of category document, including GloVe~\cite{Glove} vectors and VGSE~\cite{VGSE}.
Then, typical embeddings-based methods~\cite{SJE, APN} and generative methods~\cite{ CVPR_2019_f_vaegan_d2, LsrGAN} in attribute-based ZSL are also considered in our comparison. We train these models with category documents rather than attributes as semantic embeddings.
Besides, we compare our MADS with document-based ZSL methods~\cite{GAZSL, CIZSL}.
I2DFormer~\cite{I2DFormer} and I2MVFormer~\cite{I2MVFormer} are the recent SOTA methods. 
Moreover, we also compare methods using document embeddings from language models, \textit{e.g.}, TF-IDF~\cite{TF-IDF}, GloVe~\cite{Glove}, LongFormer~\cite{LongFormer}, and, MPNet~\cite{MPNet}, as unsupervised semantic embeddings.
For fair comparisons, we use the same image encoder and text encoder. 



{
\renewcommand{\arraystretch}{1.02} 
\begin{table*}[t]
    \caption{
        Comparison with SOTA ZSL Methods in Documents as Auxiliary Information. The Best and the Second-Best Results Are Marked in \textbf{Bold} and \underline{Underline}, Respectively. \textbf{AUX} Denotes Auxiliary Information. CLSN and IMG Are Class Names and Images. 
        }
    \vskip -0.04in
    \centering
    \small
	\resizebox{0.89\linewidth}{!}{%
	\begin{tabular}{ l | c | c c c | c c c | c c c | c c c }
			\toprule
			& & \multicolumn{3}{c}{\textbf{Zero-Shot Learning}} & \multicolumn{9}{c}{\textbf{Generalized Zero-Shot Learning}} \\
			\cmidrule(lr){3-5} \cmidrule(lr){6-14}
			\textbf{Model}	&  \textbf{AUX} & \makecell[c]{\textbf{AWA2}} & \makecell[c]{\textbf{CUB}} & \makecell[c]{\textbf{FLO}} & \multicolumn{3}{c}{\textbf{AWA2}} & \multicolumn{3}{c}{\textbf{CUB}} & \multicolumn{3}{c}{\textbf{FLO}}  \\
			\cmidrule(lr){3-3} \cmidrule(lr){4-4} \cmidrule(lr){5-5} \cmidrule(lr){6-8} \cmidrule(lr){9-11} \cmidrule(lr){12-14}
			 &  & \makecell[c]{\textbf{T1}} &  \makecell[c]{\textbf{T1}} & \makecell[c]{\textbf{T1}}& \textbf{U} & \textbf{S} & \makecell[c]{\textbf{H}} & \textbf{U} & \textbf{S} & \makecell[c]{\textbf{H}} & \textbf{U} & \textbf{S} & \makecell[c]{\textbf{H}} \\
			
            \midrule
            {GloVe}~\cite{Glove} & {CLSN} & 52.1 & 20.4 & 21.6 & 42.1 & 75.3 & 54.0 & 16.2 & 43.6 & 23.6 & 14.4 & 88.3 & 24.8 \\
            {VGSE}~\cite{VGSE} & {CLSN$+$IMG} & 69.6 & 37.1 & - & 56.9 & 82.8 & 67.4 & 27.6 & \underline{70.6} & 39.7 & - & - & -\\ 
            {GAZSL}~\cite{GAZSL} & Wiki & 63.7 & 43.7 & 20.9 & 22.2 & \textbf{90.8} & 35.6  & 15.9 & 50.4 & 24.1 & 8.4 & \underline{97.3} & 15.4 \\
            {f-VAEGAN-D2}~\cite{CVPR_2019_f_vaegan_d2} & Wiki & 70.7 & 31.8 & 32.1 & 65.7 & 69.5 & 67.6 & 23.9 & 55.7 & 33.5 & 25.0 & \textbf{99.0} & 39.9 \\
            {CIZSL}~\cite{CIZSL} & Wiki & 67.8 & 44.6 & - & - & - & - & - & - & -& - & - & - \\
            LsrGAN~\cite{LsrGAN} & Wiki & 66.4 & 45.2 & - & - & - & - & 30.2 & 57.1 & 39.5 & - & - & -   \\
            {SJE}~\cite{SJE} & Wiki & 56.6 & 27.1 & 13.1 & 41.3 & 83.4 & 55.3 & 14.4 & 51.6 & 22.5 & 4.6 & 93.2 & 8.7 \\
            {APN}~\cite{APN} & Wiki & 73.8 & 20.7 & 15.2 & 57.6 & 84.6 & 68.5 & 19.6 & 32.6 & 24.5 & 12.8 & 39.4 & 19.3 \\
            {TF-IDF}~\cite{TF-IDF} & {Wiki} & 46.4 & 39.9 & 34.0 & 29.6 & \underline{87.6} & 44.2 & 29.0 & 52.1 & 37.3 & 28.9 & 94.8 & 44.3 \\ 
            {GloVe}~\cite{Glove} & {Wiki} & 61.6 & 29.0 & 25.8 & 49.5 & 78.1 & 60.6 & 23.8 & {62.6} & 34.5 & 14.7 & 91.0 & 25.3\\
            {LongFormer}~\cite{LongFormer} & {Wiki} & 44.2 & 22.6 & 8.8 & 41.6 & 81.8 & 55.2 & 19.9 & 41.0 & 26.8 & 8.8 & 89.8 & 16.0 \\
            {MPNet}~\cite{MPNet} & {Wiki} & 61.8 & 25.8 & 26.3 & 58.0 & 76.4 & 66.0 & 20.6 & 44.3 & 28.2 & 22.2 & {96.7} & 36.1 \\
            
            \midrule
            
            {I2DFormer \cite{I2DFormer}}  &  {Wiki$+$LLM}  & 
            77.3 & 47.0 & 43.0 & 
            68.6 & 77.4 & 72.7 & 
            38.5 & 59.3 & 46.7 & 
            40.4 & 80.1 & 53.8 \\  

            {I2MVFormer \cite{I2MVFormer}}& {Wiki$+$LLM} & 
            \underline{79.6} &\underline{51.1} & \underline{46.2} & 
            \underline{75.7} & 79.6 & \underline{77.6} & 
            \underline{42.5} & 59.9 & \underline{49.7} & 
            \underline{41.6} & {91.0} & \underline{57.1} \\
            
            \highlight{{\textbf{MADS}\text{ (Ours)}}}
            & \highlight{{Wiki$+$LLM}}  & 
            \highlight{\textbf{86.7}}  & \highlight{\textbf{57.6}} & \highlight{\textbf{54.3}} & 
            \highlight{\textbf{81.5}} & \highlight{86.3} & \highlight{\textbf{83.9}} & 
            \highlight{\textbf{47.9}} & \highlight{\textbf{71.0}} & \highlight{\textbf{57.2}}  & 
            \highlight{\textbf{52.0}} & \highlight{{97.2}} & \highlight{\textbf{67.8}} \\

            \bottomrule
	\end{tabular}
}
	\label{tab:main_exp}
    \vskip -0.15in
\end{table*}
}

\textbf{Results.}
In Table~\ref{tab:main_exp}, we see that our MADS consistently outperforms SOTA methods across all metrics (T1 and H) in three benchmark datasets. 
Compared to word-embedding methods, our MADS performs better on two fine-grained datasets, \textit{i.e.}, CUB and FLO.
This verifies that classification with class-specific visual descriptions provides more discriminative information than only category names.
Besides, typical attribute-based methods (\textit{i.e.}, SJE, APN, f-VAEGAN-D2, and LsrGAN) show worse results than our MADS.
This is because these methods do not filter out non-visual noises in documents, which makes it hard to transfer knowledge.
Moreover, similar results are observed for document-based methods (\textit{i.e.}, GAZSL, CIZSL, I2DFormer, and I2MVFormer) and document embeddings methods (\textit{i.e.}, GloVe, LongFormer and MPNeT).
Compared to I2MVFormer, we surpass the SOTA method with improvements of 7.1\% and 6.3\% on AWA2, 6.5\% and 7.5\% on CUB, and 8.1\% and 10.7\% on FLO in ZSL and GZSL settings, respectively.
We attribute this high-performance gain to two reasons. 
First, we collect detailed multi-attribute documents instead of noisy and coupled documents in previous methods as auxiliary information, offering a better theoretical foundation for knowledge transfer.
Second, our MADS network achieves accurate semantic alignment between visual words of each view and salient regions by explicit attention penalization.



{

    \begin{table}[t]
    \centering
    \caption{Results of Methods with Multi-Attribute Documents (MAD) as Input. \textbf{DOC} is Category Documents for Auxiliary Information.}
    \vskip -0.04in
    \label{tab:SOTA_with_MAD}
        \resizebox{0.99\linewidth}{!}
        {
        \begin{tabular}{l | c | ccc | ccc}
        \toprule
        \multirow{2}{*}{\textbf{Model}} & \multirow{2}{*}{\textbf{DOC}} & \multicolumn{3}{c}{\textbf{CUB}} & \multicolumn{3}{c}{\textbf{FLO}}\\
        \cmidrule(lr){3-5} \cmidrule(lr){6-8} 
        & & \textbf{U} & \textbf{S} & \textbf{H} & \textbf{U} & \textbf{S} & \textbf{H}\\
        \midrule 
        I2DFormer~\cite{I2DFormer} & MAD & 40.7 & 61.0 & 48.8 & 42.7 & 94.0 & 58.7 \\
        I2MVFormer~\cite{I2MVFormer} & MAD & 41.9 & 63.6 & 50.5 & 42.6 & \textbf{98.1} & 59.4 \\
        \highlight{\textbf{MADS} (Ours)} & \highlight{MAD}  & \highlight{\textbf{47.9}} & \highlight{\textbf{71.0}}  & \highlight{\textbf{57.2}} & \highlight{\textbf{52.0}} & \highlight{97.2} &  \highlight{\textbf{67.8}} \\
        \bottomrule
        \end{tabular}
        }
        \vskip -0.1in
    \end{table}
}

\textbf{SOTA Methods with Multi-Attribute Document as Input.}  
To verify the effectiveness of our MADS in extracting transferable knowledge from multi-attribute documents, we also train the SOTA models (\textit{i.e.}, I2DFormer~\cite{I2DFormer} and I2MVFormer~\cite{I2MVFormer}) with the same inputs on CUB and FLO datasets.
For I2DFormer, we integrate multi-attribute documents into one paragraph and input it.
For I2MVFormer, we feed each paragraph as a view document into the model.
As shown in Table~\ref{tab:SOTA_with_MAD}, our MADS achieves STOA results with improvements of 6.7\% on CUB and 8.4\% and FLO in GZSL settings, which is consistent with the discussion in Section~\ref{sec: obtain_semantic}.
This is because previous methods fail to capture the semantic interaction and ignore the varied contributions of each attribute paragraph for image recognition.
Our MADS considers both of them, yielding the best result.


{
    \setlength{\tabcolsep}{4.pt}
    \renewcommand{\arraystretch}{1.02} 
    \begin{table}[t]
    \centering
    \caption{Performance Improvement of Previous Methods by Removing Noisy Descriptions (Algorithm~\ref{alg: collect_document}) and Adding Focus Loss.
    Gains Compared With the Baseline ($\Delta$) Are Shown in \blue{Blue}. 
    The Best Result Within a Method is \underline{Underline}.}
    \vskip -0.04in
    \label{tab: improve_previous_method}
        \resizebox{0.95\linewidth}{!}
        {
        \begin{tabular}{l | ccc | ccc | ccc}
        \toprule
        \multirow{2}{*}{\textbf{Model}} & \multicolumn{3}{c}{\textbf{AWA2}} & \multicolumn{3}{c}{\textbf{CUB}} & \multicolumn{3}{c}{\textbf{FLO}}\\
        \cmidrule(lr){2-4} \cmidrule(lr){5-7} \cmidrule(lr){8-10} 
        & \textbf{U} & \textbf{S} & \textbf{H} & \textbf{U} & \textbf{S} & \textbf{H} & \textbf{U} &  \textbf{S} & \textbf{H}\\
        \midrule 
        I2DFormer\cite{I2DFormer} & 68.6 & 77.4 & 72.7 &  38.5 & 59.3 & 46.7 & 40.4 & 80.1 & 53.8 \\
        +Alg.~\ref{alg: collect_document} & 72.9 & 85.2 & 78.6 & 40.7 & 61.0 & 48.8 & 42.7 & \underline{94.0} & 58.7 \\ 
        +Alg.~\ref{alg: collect_document}+$\mathcal{L}_{focus}$ & \underline{75.3} & \underline{\textbf{88.7}} & \underline{81.5} & \underline{41.6} & \underline{61.7} & \underline{49.7} & \underline{47.0} & 87.9 & \underline{61.2} \\
        $\Delta$ & \blue{+6.7} & \blue{+11.3} & \blue{+8.8} & \blue{+3.1} & \blue{+2.4} & \blue{+3.0} & \blue{+6.6} & \blue{+7.8} & \blue{+7.4}  \\
        \midrule
        I2MVFormer\cite{I2MVFormer} & 75.7 & 79.6 &  77.6 & 42.5 & 59.9 & 49.7 & 41.6 & 91.0 & 57.1 \\
        +Alg.~\ref{alg: collect_document} & 76.3 & 83.9 & 79.9 & 43.4 & 61.4 & 50.9 & 44.8 & \underline{94.0} & 60.6 \\ 
        +Alg.~\ref{alg: collect_document}+$\mathcal{L}_{focus}$ & \underline{77.1} & \underline{87.6} & \underline{82.0} & \underline{44.4} & \underline{63.2} & \underline{52.2} & \underline{48.5} & 87.7 & \underline{62.5} \\
        $\Delta$ &  \blue{+1.4} & \blue{+8.0} & \blue{+4.4} & \blue{+1.9} & \blue{+3.3} & \blue{+2.5} & \blue{+6.9} & {-3.3} & \blue{+5.4} \\
        \midrule
        
        \highlight{\textbf{MADS} (Ours)} & \highlight{\textbf{81.5}} & \highlight{86.3} & \highlight{\textbf{83.9}} & 
            \highlight{\textbf{47.9}} & \highlight{\textbf{71.0}} & \highlight{\textbf{57.2}}  & 
            \highlight{\textbf{52.0}} & \highlight{\textbf{97.2}} & \highlight{\textbf{67.8}} \\
        \bottomrule
        \end{tabular}
        }
        \vskip -0.09in
    \end{table}
}

{
\begin{figure}[tb]
\begin{center}
\centerline{\includegraphics[width=\linewidth]{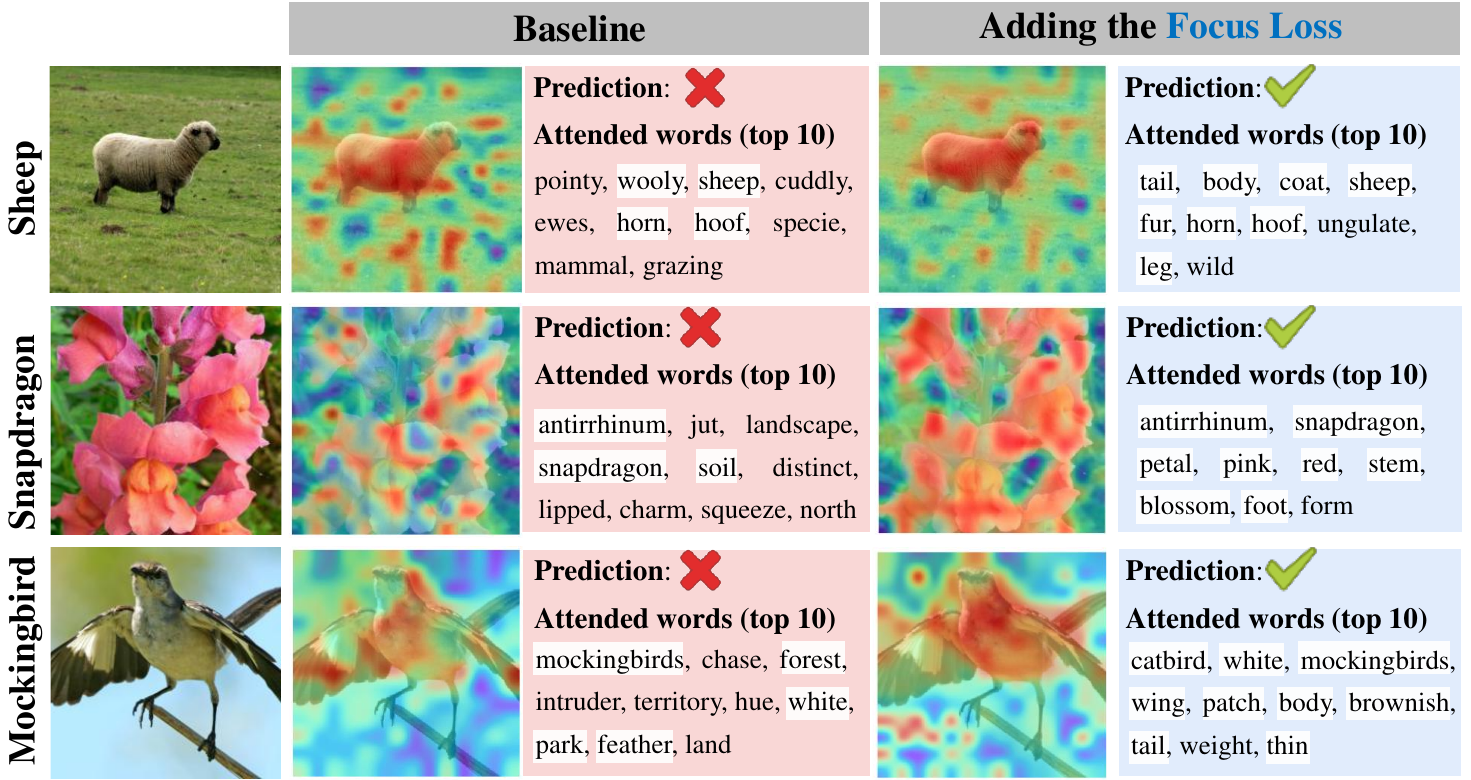}}
\vskip -0.07in
\caption{
Visualization of salient image regions and most attended words in attention mechanism.
Although both the baseline and the model with the focus loss extract the salient visual regions, the latter pays more attention to visual words (shown with the white area) that are helpful for ZSL tasks.
}
\label{fig: Baseline_improve}
\end{center}
\vskip -0.2in
\end{figure}
}

\subsection{Performance Improvement of Previous Methods}
\label{exp: improve_previous_methods}

In this section, we show that our proposed document collection algorithm and model-agnostic focus loss can consistently improve the performance of previous methods.

\textbf{Performance Improvement by Removing Noisy Descriptions in Documents.}
Unlike previous methods that remove noises by model design, \textit{e.g.}, regularized loss or additional FC layer, we introduce Algorithm~\ref{alg: collect_document} to remove non-visual noisy sentences and enrich visual descriptions in documents by LLMs.
As shown in Table~\ref{tab: improve_previous_method}, we see that our collected documents are also helpful for previous methods, improving them consistently on three datasets.
This verifies the effectiveness of noise suppression in our algorithm at the document collection stage.
Notably, we integrate our documents into one paragraph and rewrite them in various styles by LLMs, which are then inputted for I2MVFormer~\cite{I2MVFormer}.
This provides a better performance for I2MVFormer (50.9\% in CUB and 60.6\% in FLO) than treating each attribute paragraph as a view (50.5\% in CUB and 59.4\% in FLO) but still with information coupling.

\textbf{Focus Loss Makes the Model Pay More Attention to Visual Words.}
Since the focus loss only penalizes attention mechanisms in the model, we can easily add this loss to previous methods. They also use attention mechanisms to extract discriminative information from documents.
As shown in Table~\ref{tab: improve_previous_method}, we see performance improvements by 3.0\% to 8.8\% in I2DFormer~\cite{I2DFormer} and by 2.5\% to 5.4\% in I2MVFormer~\cite{I2MVFormer}.
To further verify the reason for improvement by focus loss, we visualize the salient image regions and the most attended words (top 10) in the model before and after adding the focus loss in Figure~\ref{fig: Baseline_improve}.
We see that the baseline model may incorrectly relate salient image regions with non-visual words, \textit{e.g.}, \texttt{specie} and \texttt{grazing} in Sheep, \texttt{jut} and \texttt{distinct} in Snapdragon, and \texttt{chase} and \texttt{intruder} in Mockingbird.
After adding the focus loss, the model accurately attends visual words, especially about shape (\texttt{patch} and \texttt{thin} in Mockingbird), object parts (\texttt{body} and \texttt{leg} in Sheep), and color (\texttt{pink} and \texttt{red} in Snapdragon) of the category.
This validates the effectiveness of explicit constrain on visual information in the attention mechanism by our focus loss.


{
\begin{figure}[t]
\begin{center}
\centerline{\includegraphics[width=\linewidth]{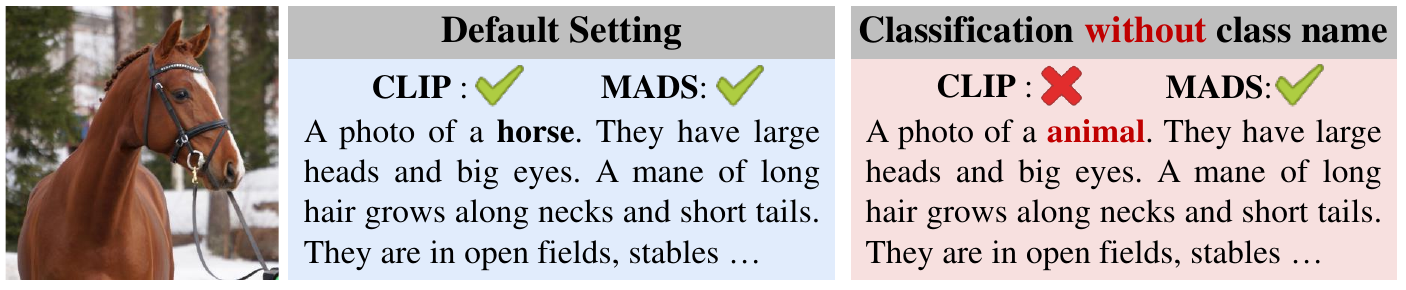}}
\vskip -0.07in
\caption{
{
Illustration of two settings. Default 
setting: We use the class name and corresponding descriptions to classify.
Classification without class name: We replace the name with the dataset domain to remove priors on class names.
}
}
\label{fig:clip_without_name}
\end{center}
\vskip -0.17in
\end{figure}
}

{
    \setlength{\tabcolsep}{4.5 pt}

    \begin{table}[t]
    \centering
    \caption{Compared to CLIP-based Description Methods. 
    We Consider Two Settings: Default (w/ class) and Classification without Class Name (w/o class). The best Results Are in \textbf{Bold}.
    Highest and Lowest Results Changes ($\Delta$) Are Shown in \textcolor{red}{Red} and \blue{Blue}.
    }
    \vskip -0.04in
    \label{tab:compare_clip}
    \resizebox{0.85\linewidth}{!}
        {
        \begin{tabular}{l | ccc | ccc}
        \toprule
        \multirow{2}{*}{\textbf{Model}} &  \multicolumn{3}{c}{\textbf{CUB}} & \multicolumn{3}{c}{\textbf{FLO}} \\
        \cmidrule(lr){2-4} \cmidrule(lr){5-7}
        & w/ class & w/o class & $\Delta$ & w/ class & w/o class & $\Delta$  \\
        \midrule
        \multicolumn{7}{c}{\textit{\textbf{\ccol{CLIP-Based Description Methods}}}} \\
        \midrule
        {CLIP}~\cite{CLIP} & 54.1 & - & - & 66.8 & - & -  \\
        \midrule
        DCLIP~\cite{DCLIP} & 57.1 & 4.1 & \textcolor{red}{-53.0} & 70.7 & 7.5 & \textcolor{red}{-63.2}  \\
        CuPL~\cite{CuPL} & 57.2 & 16.1 & -41.1 & \textbf{72.5} & 29.4  & -43.1 \\
        {our document}  & \textbf{58.3} & 14.1 & -44.2 & 71.7 & 18.2 & -53.5 \\
        
        \midrule
        \multicolumn{7}{c}{\textit{\textbf{\ccol{Specific-Domain Description Methods}}}} \\
        \midrule
        I2DFormer~\cite{I2DFormer} & 46.7 & 38.1 & -8.6 & 53.8 & 41.7 & -12.1\\
        I2MVFormer~\cite{I2MVFormer} & 49.7 & 35.9 & -13.8 & 57.1 & 44.3 & -12.8 \\
        \highlight{\textbf{MADS}}  & \highlight{57.2} & \highlight{\textbf{50.4}} & \highlight{\blue{-6.8}} & \highlight{67.8} & \highlight{\textbf{58.2}} & \highlight{\blue{-9.6}} \\
        \bottomrule
        \end{tabular}}
        \vskip -0.15in
    \end{table}
}

\subsection{Comparison with CLIP-based Description Methods}
\label{sec:compare_with_clip}
In this Section, we compare our MADS with CLIP-based description methods (\textit{i.e.}, DCLIP~\cite{DCLIP} and CuPL~\cite{CuPL}) in two settings. 
As shown in Figure~\ref{fig:clip_without_name}, except for the default classification settings, we consider a setting lacking the prior information of class name, \textit{i.e.}, a category newly emerges and its name not seen in the training, solely described by visual descriptions.
To remove priors of class names in CLIP, we use ``\texttt{a photo of a \{type\}. \{descriptions\}.}'' as template in this situation, where \{\texttt{type}\} is the dataset domain. 
In Table~\ref{tab:compare_clip}, we show the results of CLIP-based and specific-domain description methods under the two settings.
The performance of CLIP increases with the help of class-specific descriptions under the default setting, where our document can also improve CLIP by 4.2\% in CUB and 4.9\% in FLO.
However, there is a notable performance drop by 53.0\% to 63.2\% in DCLIP~\cite{DCLIP} and by 41.1\% to 43.1\% in CuPL~\cite{CuPL} under the setting without class names.  
This verifies that the ability of CLIP-based description methods to recognize categories relies heavily on the prior information of class names in CLIP.
In contrast, MADS outperforms them and demonstrates superior generalization for fine-grained recognition under the setting without class name priors. 
Compared to SOTA specific-domain description methods~\cite{I2DFormer, I2MVFormer}, our MADS also has minimal performance degradation in the two settings.
This is because we relate associations between words that describe visual details about the category rather than only class name and their corresponding image regions.

{
    \setlength{\tabcolsep}{4pt}

    \begin{table}[t]
    \centering
    \caption{Ablation of Processes in Document Collection.}
    \vskip -0.05in
    \label{tab: Ablation_Prompt}
        \resizebox{0.8\linewidth}{!}
        {
        \begin{tabular}{l | ccc | ccc}
        \toprule
        \multirow{2}{*}{\textbf{Model}} &  \multicolumn{3}{c}{\textbf{CUB}} & \multicolumn{3}{c}{\textbf{FLO}}\\
        \cmidrule(lr){2-4} \cmidrule(lr){5-7} 
        & \textbf{U} & \textbf{S} & \textbf{H} & \textbf{U} & \textbf{S} & \textbf{H}\\

        \midrule
        \multicolumn{7}{c}{\textit{\textbf{\ccol{Generate Visual Attribute Views}}}} \\
        \midrule
        \texttt{a)\;\;}w/ directly remove noises & 40.3 & 65.6 & 49.9 & 43.1 & \textbf{97.7} & 59.8 \\
        \texttt{b)\;\;}w/ all attribute views &  38.8 & 68.1 & 49.5 & 44.1 & 87.1 & 58.6 \\
        
        \midrule
        \multicolumn{7}{c}{\textit{\textbf{\ccol{Divide Document by Visual Attributes Views}}}} \\
        \midrule
        \texttt{c)\;\;}w/ definition document & 44.5 & 58.6 & 50.6 & 47.1 & 86.1 & 60.9 \\
        \texttt{d)\;\;}w/ one paragraph & 44.6 & 66.2 & 53.3 & 46.7 & 95.7 & 62.8 \\
    
        \midrule
        \multicolumn{7}{c}{\textit{\textbf{\ccol{Enrich Less-Described Attribute Document}}}} \\
        \midrule
        
         \texttt{e)\;\;}w/o enrich document & 40.3 & 65.6 & 49.9 & 44.0 & 93.8 & 59.9 \\
         \texttt{f)\;\;}w/o constraints on visual detail & 38.7 & 67.1 & 49.1 & 47.5 & 89.3 & 62.0 \\
         \texttt{g)\;\;}w/ comparable descriptions & 39.7 & 68.2 & 50.2 & 45.7 & 92.0 & 61.1 \\
          \texttt{h)\;\;}w/ directly enrich document & 43.0 & 68.1 & 52.7 & 47.1 & 86.1 & 60.9 \\
         
        \midrule
         \highlight{\textbf{MADS} (multi-attribute document)} & \highlight{\textbf{47.9}} & \highlight{\textbf{71.0}} & \highlight{\textbf{57.2}} & \highlight{\textbf{52.0}} & \highlight{{97.2}} & \highlight{\textbf{67.8}} \\
        
        \bottomrule
        \end{tabular}
        }
        \vskip -0.1in
    \end{table}
}

{

    \setlength{\tabcolsep}{4pt}
    \renewcommand{\arraystretch}{1.02} 
    
    \begin{table}[t]
    \centering
    \caption{Ablation on Different LLMs. The \textbf{SRDD} Denotes the Success Rate of Dividing Documents Based on Visual Attributes.}
    \vskip -0.05in
    \label{tab:Ablation_on_LLM}
        \resizebox{0.93\linewidth}{!}
        {
        \begin{tabular}{l | ccccc}
        \toprule
        \textbf{Model} & {GPT3}~\cite{gpt3} & {Llama2}~\cite{llama2} & {PaLM2}~\cite{PaLM} & {ChatGPT~\cite{ChatGPT}} & {GPT-4~\cite{GPT4_Tech}} 
        \\
        \midrule
        \textbf{SRDD} (\%) & 10 & 46 & 60 & 88 & \textbf{100} \\
        \bottomrule
        \end{tabular}
        }
        \vskip -0.15in
    \end{table}
}

\subsection{Ablation on Document Collection.}
In this section, we study the effect of key components of document collection.
Table~\ref{tab: Ablation_Prompt} shows the ablation results of three processes.
\textbf{(1) Ablation of $p_{view}$}: By asking LLMs to remove noisy descriptions directly, we see a result drop in row a). This process incorrectly removes visual sentences as noises, which may cause the document to lack sufficient semantics. In row b), we query LLMs to generate all potential views for defining the category instead of only visual attribute views. The same performance degradation appears due to the introduction of non-visual views, resulting in invalid noise suppression.
\textbf{(2) Ablation of $p_{divide}$}: In row c), we use the original documents from the encyclopedia without any processing as auxiliary information. The performance drop appears due to the non-visual noise and information coupling in documents. We also integrate multi-attribute documents into one paragraph similar to I2DFormer~\cite{I2DFormer} to train the model. This achieves a suboptimal result compared to full MADS, demonstrating the effectiveness of inputting each attribute paragraph to the model with information decoupling.
\textbf{(3) Ablation of $p_{enrich}$}: Without this process, there is a performance drop in row e) due to the lack of sufficient semantics to train. Without constraints on visual details in the prompt $p_{enrich}$, there may be non-visual noise in descriptions. Similar noises appear when generating comparable descriptions for each attribute view. Therefore, the performance drops in rows f) and g). We also try to enrich the collected documents instead of the divided documents. In row h), a performance degradation appears, which verifies that enriching divided documents provides more detailed descriptions and achieves better performance.
\textbf{(4) Ablation on Different LLMs}: In Table \ref{tab:Ablation_on_LLM}, we test different LLMs (\textit{i.e.}, GPT3 \cite{gpt3}, Llama2 \cite{llama2}, PaLM2 \cite{PaLM}, ChatGPT~\cite{ChatGPT}, and GPT-4~\cite{GPT4_Tech}) to collect multi-attribute documents in a zero-shot manner. GPT-4 perfectly completes the division of the document. However, other LLMs may fail due to incorrectly generated formats or the inability to divide documents into paragraphs. We leverage the GPT-4 for multi-attribute document collection and will explore other LLMs in a few-shot way in future work.





{
    \setlength{\tabcolsep}{4.5pt}

    \begin{table}[t]
    \centering
    \caption{Ablation of Key Components in Model Learning.}
    \vskip -0.05in
    \label{tab: Ablation_Module}
        \resizebox{0.81\linewidth}{!}
        {
        \begin{tabular}{l | ccc | ccc}
        \toprule
        \multirow{2}{*}{\textbf{Model}} &  \multicolumn{3}{c}{\textbf{CUB}} & \multicolumn{3}{c}{\textbf{FLO}}\\
        \cmidrule(lr){2-4} \cmidrule(lr){5-7} 
        & \textbf{U} & \textbf{S} & \textbf{H} & \textbf{U} & \textbf{S} & \textbf{H}\\
        \midrule

        \multicolumn{7}{c}{\textit{\textbf{\ccol{Loss Functions}}}} \\
        \midrule
         \texttt{a)\;\;}w/o $\mathcal{L}_{local}$ & 41.2  & 66.0 & 50.7 & 42.2  & 93.0 & 58.0 \\
         \texttt{b)\;\;}w/o $\mathcal{L}_{focus}$ & 45.3 & 63.2 & 52.8 &  45.1 & 97.6 & 61.7 \\
         \texttt{c)\;\;}w/o $\mathcal{L}_{local}$ + $\mathcal{L}_{focus}$ & 41.1  & 60.8  & 49.1 & 40.2 & 97.0 & 56.8\\

        \midrule 
        \multicolumn{7}{c}{\textit{\textbf{\ccol{Modules in MADS Network}}}} \\
        \midrule
        \texttt{d)\;\;}w/o semantic perceiver & 41.9 & 63.6 & 50.5 & 47.5 & 95.5 & 63.5 \\
        \texttt{e)\;\;}w/o weighted sum in Eq. \ref{eq:weighed_sum} & 43.5 & 64.9 & 52.1 & 47.8 & \textbf{99.0} & 64.4 \\
        \texttt{f)\;\;}w/o residual connect in Eq. \ref{eq:res_connect} & 43.7 & 63.4 & 51.8 & 47.5 & 89.3 & 62.0 \\
        \texttt{g)\;\;}w/ FILIP \cite{FILIP}  & 39.7 & 61.5 & 48.3 & 40.7 & 94.2 & 56.9\\
        \texttt{h)\;\;}w/o aggregate module & 41.9 & 66.4 & 51.4  & 43.1 & 97.7 & 59.8 \\
        
        \midrule
        \multicolumn{7}{c}{\textit{\textbf{\ccol{Design of Focus Loss}}}} \\
        \midrule
         \texttt{i)\;\;}w/ only visual words & 42.1 & 63.4 & 50.6  &  44.1 & 87.1 & 58.6 \\
         \texttt{j)\;\;}w/ focus on non-visual words & 44.2 & 55.6 & 49.2 & 39.4 & 97.2 & 56.1 \\
         \texttt{k)\;\;}w/ VG label \cite{VG_dataset} & 46.5 & 68.8 & 55.5 & 50.0 & 96.4 & 65.8 \\
         \texttt{l)\;\;}w/ KL loss & 47.1 & 63.2 & 54.0 & 48.0 & 92.3 & 63.1 \\

        \midrule
        
        \highlight{\textbf{MADS} (full model)} &  \highlight{\textbf{47.9}} & \highlight{\textbf{71.0}}  & \highlight{\textbf{57.2}} & \highlight{\textbf{52.0}} & \highlight{97.2} &  \highlight{\textbf{67.8}} \\
         
        \bottomrule
        \end{tabular}
        }
        \vskip -0.15in
    \end{table}
}


\subsection{Ablation over Model Architecture}
In Table \ref{tab: Ablation_Module}, we evaluate the effect of key components in our MADS network. 
\textbf{(1) Ablation of Loss Functions}: In row a), we see a performance drop without $\mathcal{L}_{local}$, which is due to the lack of fine-grained interactions between words and image regions. Performance is decreased by the absence of $\mathcal{L}_{focus}$ in row b). It shows the importance of explicitly attending to the visual information through focus loss. Row c) achieves the worst performance without $\mathcal{L}_{local}$ and $\mathcal{L}_{focus}$, further verifying the effectiveness of our losses.
\textbf{(2) Ablation of Modules in MADS}: Without the semantic perceiver to extract core semantic information and filter noises of each attribute view, there is a result decrease in row d). We see the same performance degradation in rows e) and f) without the weighted sum of Eq.~\ref{eq:weighed_sum} and residual connect of Eq.~\ref{eq:res_connect}. This is due to the absence of independent view semantics, which causes information coupling again in local and global semantic embeddings. We also replace the local semantic alignment module with FILIP~\cite{FILIP}, a parameter-free mechanism for fine-grained interactions. A significant performance decrease in row g) verifies the importance of deep feature fusion by our local module. Without the aggregate module, the performance decreases in row h) due to the lack of semantic interaction between different attribute views. 
\textbf{(3) Ablation of Design of Focus Loss}: 
\label{sec: ablate_focus_loss}
In row i), we see a performance degradation when only visual words are used. This is because the other words offer contextual information and assign different values for visual words. When penalizing the model to focus on non-visual words, there is a significant drop in row j). It demonstrates that non-visual noise is detrimental to knowledge transfer. Replacing RAM~\cite{RAM_label} label with VG label~\cite{VG_dataset} in Eq.~\ref{eq: focus_loss}, performance is slightly decreased in row k) due to the reduction of the number of labels. We also use Kullback-Leibler (KL) loss in $\mathcal{L}_{focus}$, yielding a drop in row l). This is due to the KL loss as the distributional constraint, which does not ensure the model focuses on specific words.


{
\begin{figure*}[t]
\centering
{
	\subfloat{\includegraphics[width=0.18\linewidth]{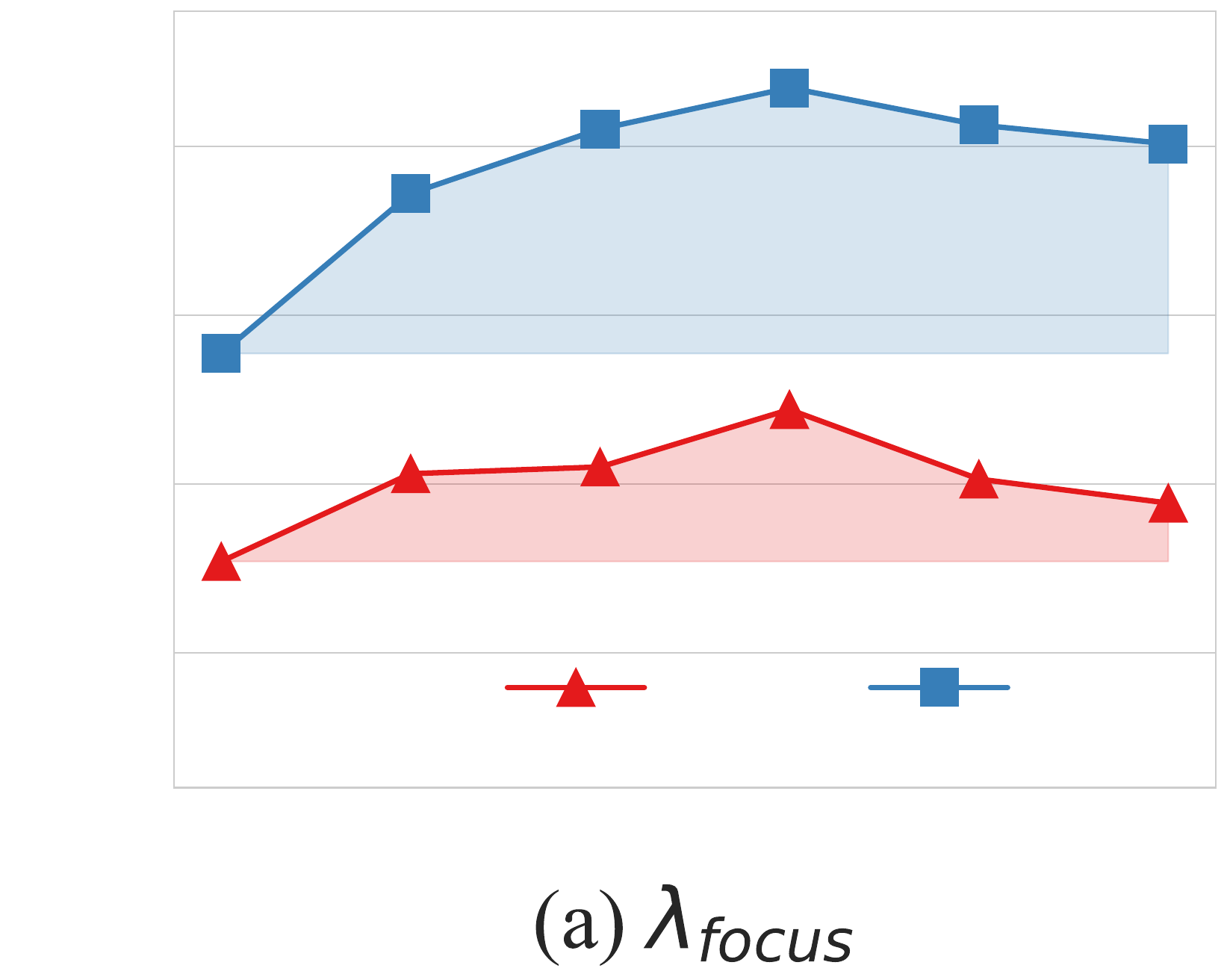}}
	\hfill
	\subfloat{\includegraphics[width=0.18\linewidth]{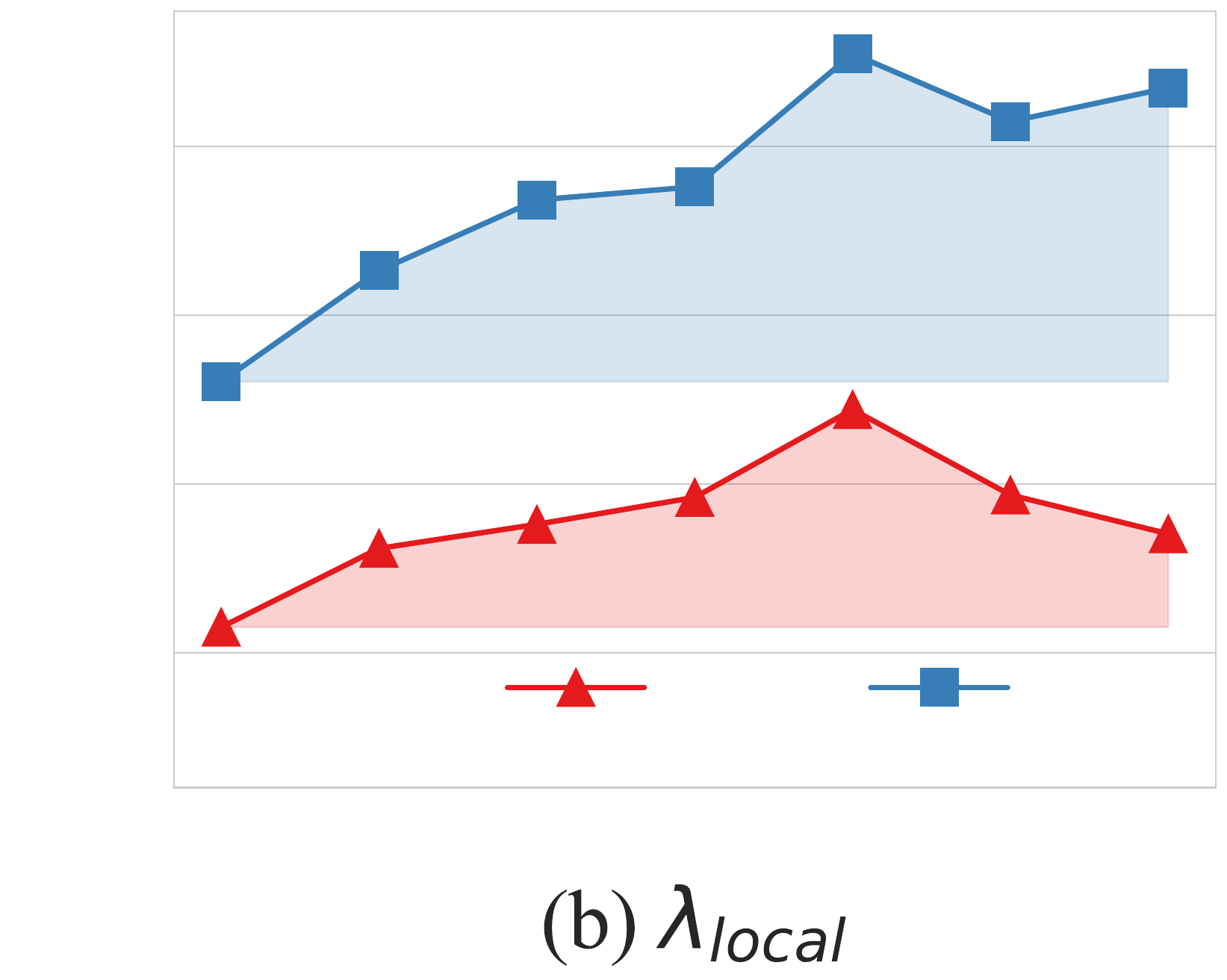}}
    \hfill
    \subfloat{\includegraphics[width=0.18\linewidth]{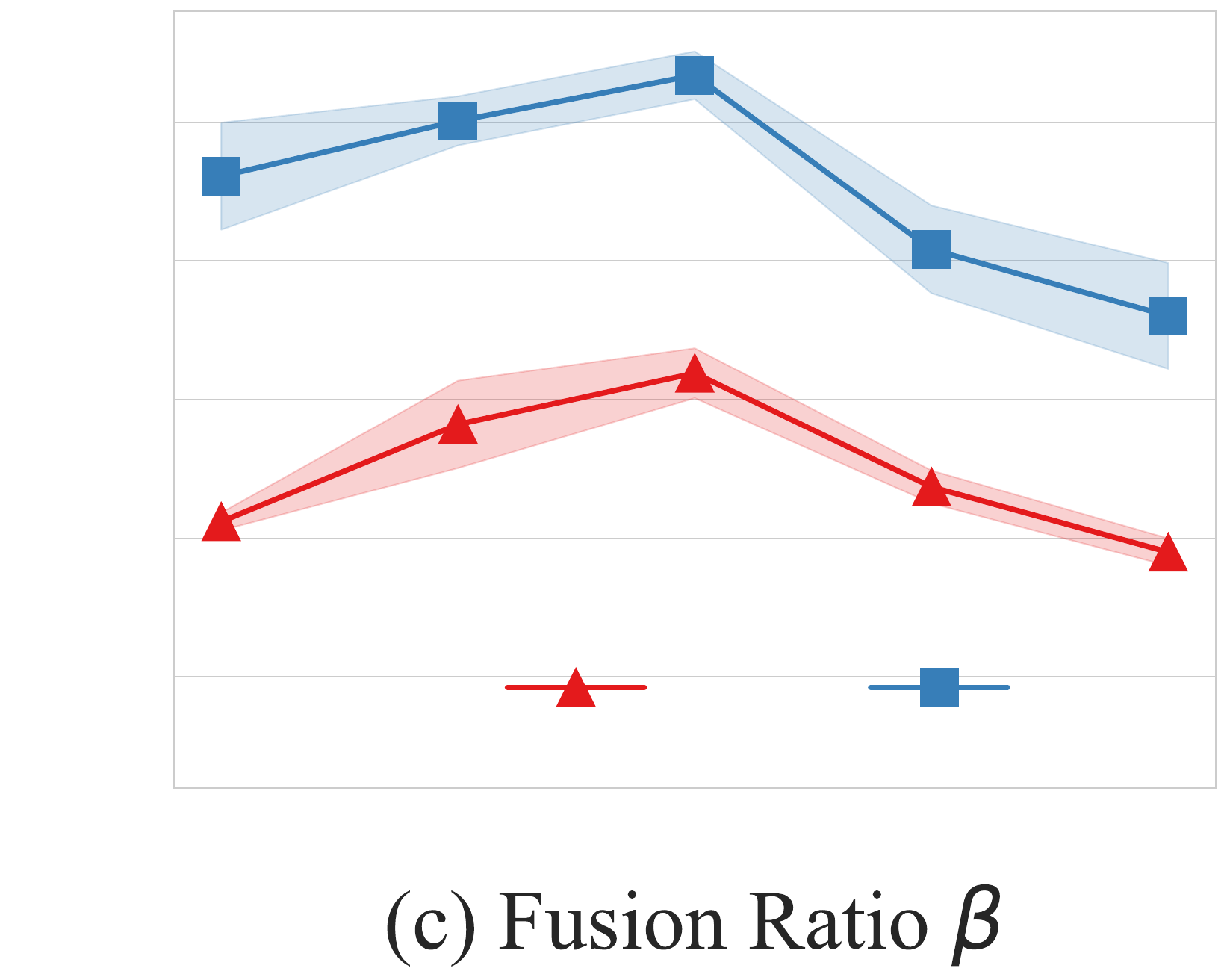}}
	\hfill
	\subfloat{\includegraphics[width=0.18\linewidth]{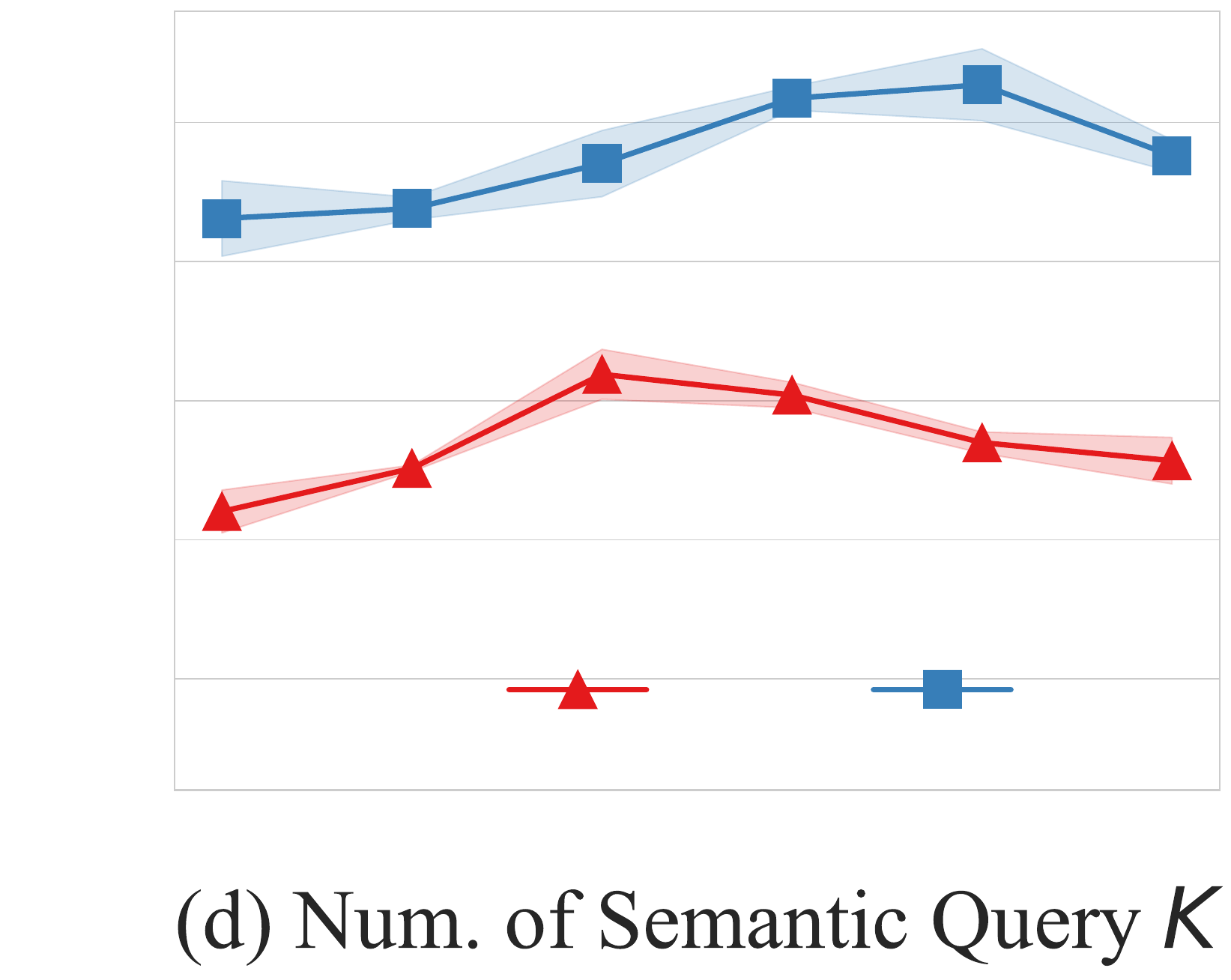}}
        \hfill
	\subfloat{\includegraphics[width=0.18\linewidth]{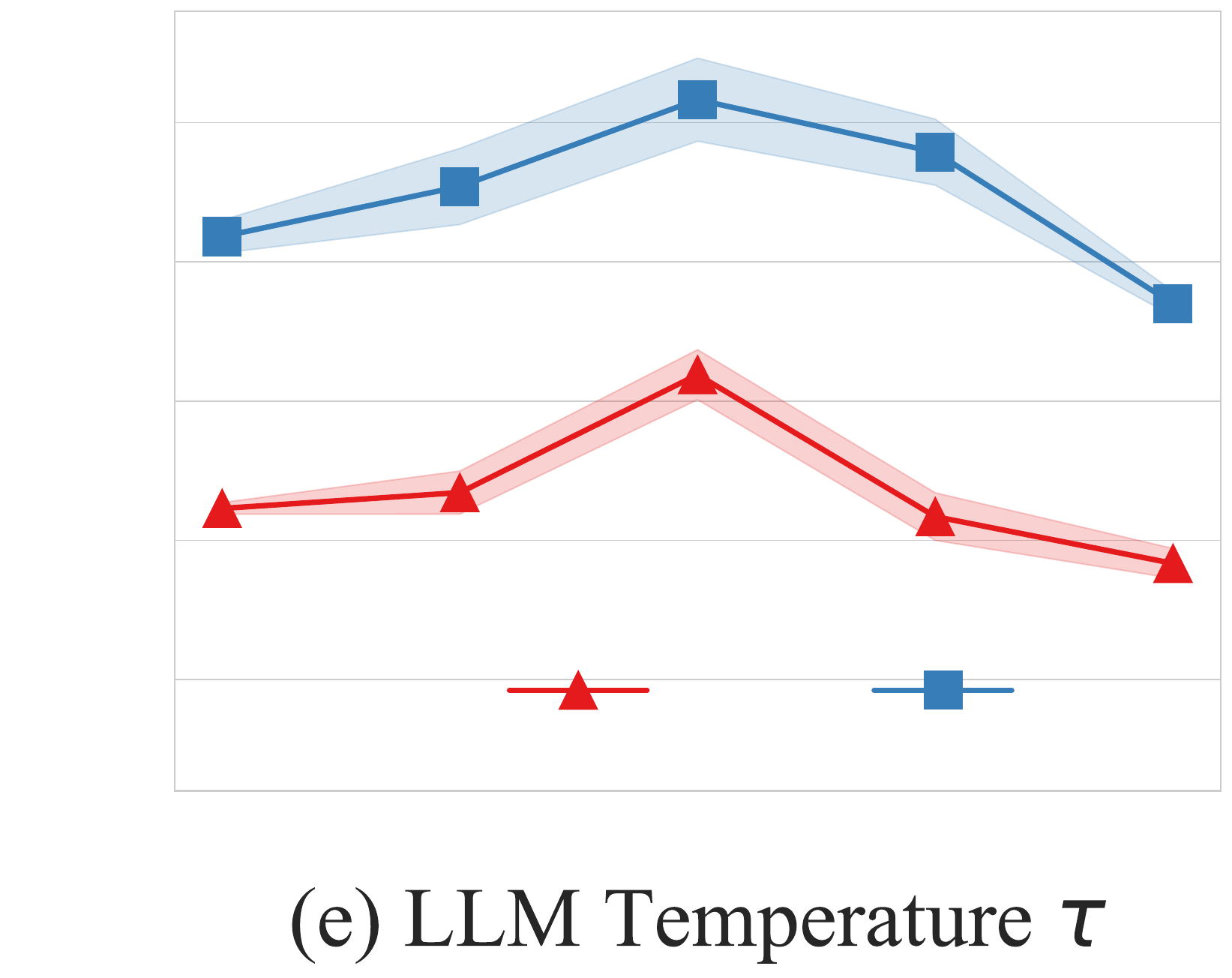}}
}
\vskip -0.1in
\caption{Effect of loss weights (a-b) and hyperparameter analysis (c-e). The shaded area in (a-b) denotes the performance improvements compared with loss weights set as 0, and in (c-e) denotes the error bars of models trained with three different documents.}
\label{fig:loss_influence}
\vskip -0.1in
\end{figure*}
}

{
\begin{figure*}[t]
\begin{center}
\centerline{\includegraphics[width=0.94\linewidth]{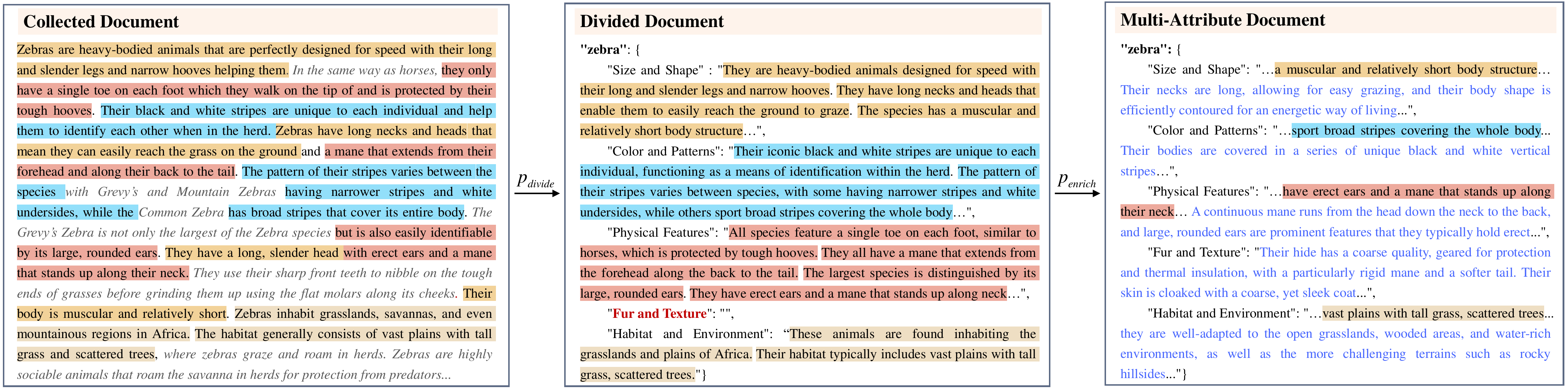}}
\vskip -0.07in
\caption{
Visualization of processes in multi-attribute document collection. Noisy descriptions in collected documents are shown in \textcolor{gray}{gray}. Visual descriptions from multiple attribute views are shown in different shaded colors. The less-described attribute view in divided documents is shown in \textbf{\red{red}}. Generated descriptions by LLMs are shown in \blue{blue}. We see that final multi-attribute documents filter out noisy descriptions and contain rich visual details about multiple views. 
}
\label{fig: MADS_document_collect}
\end{center}
\vskip -0.25in
\end{figure*}
}

\subsection{Impact of Hyperparameters}
In this section, we show the effect of hyperparameters in our MADS framework.
\textbf{(1) Effect of Loss Weights}: In Figure~\ref{fig:loss_influence}(a-b), we show the effect of loss weights in Eq.~\ref{eq: final_loss}. Notably, consistent improvements are seen as $\lambda_{focus}$ and $\lambda_{local}$ increase compared to the value set as 0, which confirms the effectiveness of our proposed losses. Adding $\lambda_{focus}$ encourages the model to pay more attention to visual words. However, high values ($\lambda_{focus} > 0.5$) might ignore the other words that are helpful for contextual understanding and cause slight performance degradation. $\mathcal{L}_{local}$ encourages fine-grained interaction, but too high might overfit the words appearing in seen classes. Therefore, we see a performance drop as the $\lambda_{local} > 0.5$.
\textbf{(2) Effect of Hyperparameters in Modules}: In Figure~\ref{fig:loss_influence}(c), we show the effect of fusion ratio $\beta$ in Eq.~\ref{eq:weighed_sum}. As $\beta$ increases, there is a progressive performance improvement. When $\beta = 0.5$, it achieves an optimal trade-off between independent and aggregated semantic embeddings in both two datasets. However, the larger $\beta$ may overly focus on aggregated information, leading to information coupling and performance decrease. In Figure~\ref{fig:loss_influence}(d), we report the influence of the number of semantic queries $K$. We see a notable performance improvement when  $K > 4$ on FLO and $K > 2$ on CUB. Since the average length of a view document in FLO is longer than in FLO, we need more queries to capture richer semantics in FLO (best $K = 8$) while less in CUB (best $K = 4$). The larger $K$ causes the model to be biased to seen classes, thus harmful to performance.
\textbf{(3) Effect of Generated Documents by LLMs}: In Figure~\ref{fig:loss_influence}(e), we show the effect of temperature $\tau$ in LLMs, a scalar to control the randomness of generated content. As $\tau$ increases, LLMs generate more varied visual descriptions, which enhances performances. However, a high temperature ($\tau > 1.0$) might generate non-visual descriptions due to increased diversity. Therefore, the performance initially improves but subsequently decreases. In Figure~\ref{fig:loss_influence}(c-e), we also show performance error bars of different documents. Our MADS achieves stable results within 1\% at the optimal result under different settings. This demonstrates the stability and effectiveness of our MADS.


{
    \setlength{\tabcolsep}{4.5pt}

    \begin{table}[t]
    \centering
    \caption{Computation Cost Analysis. 
    Performance Gains Are in \textbf{${\blue\uparrow}$}.
    }
    \vskip -0.05in
    \label{tab:compute_cost}
    \resizebox{0.75\linewidth}{!}
        {
        \begin{tabular}{l | c | ccc | l}
        \toprule
        \multirow{2}{*}{\textbf{Model}}  & \textbf{LLMs} &  \textbf{Params} & \textbf{Train}& \textbf{Inference}  & {\textbf{FLO}} \\
        & (min) &  ($\times 10^6$) &  (min)  & (ms) & (H) \\
        \midrule
        {I2DFormer} \cite{I2DFormer} & - & 0.75 & 0.72 & 4.7 &    53.8 \\
        +Alg.~\ref{alg: collect_document} + $\mathcal{L}_{focus}$ & 1.20 & 0.75 & 0.76 & 4.7 & {{61.2}}\textbf{$^{\blue{\uparrow\text{\textbf{7.4}}}}$}\\
        
        \midrule
        {I2MVFormer} \cite{I2MVFormer} & 1.02  & 1.88 &  0.50 & 5.3 &  57.1 \\
        +Alg.~\ref{alg: collect_document} + $\mathcal{L}_{focus}$ & 2.22 & 1.88 & 0.55 & 5.3 &  {62.5}\textbf{$^{\blue{\uparrow\text{\textbf{5.4}}}}$}  \\
        \midrule
        \highlight{\textbf{MADS}} & \highlight{1.20} & \highlight{1.23} & \highlight{0.67} & \highlight{6.1} & \highlight{\textbf{67.8}} \\
        \bottomrule
        \end{tabular}}
        \vskip -0.17in
    \end{table}
}

\subsection{Computation Cost Analysis}
In Table~\ref{tab:compute_cost}, we compare our MADS with SOTA methods in computation costs of document collection and model learning stages.
For document collection, we show the average annotation time for each class by LLMs.
With a slight increase in time compared to the I2MVFormer, our MADS achieves the SOTA results in three datasets.
Notably, we still save considerable human resources compared to annotating attributes.
For model learning, we reduce the number of semantic queries but increase the paragraphs of documents compared to I2MVFormer~\cite{I2MVFormer}. Hence, there is a slight increase in training and inference time but a decrease in learnable parameters.
After adding the Algorithm~\ref{alg: collect_document} and focus loss to previous methods, we see significant performance improvement and minimal impact on learnable parameters and inference time but a slight increase in training time for one epoch. 
This is because focus loss is only used at training and without introducing additional parameters.
Generally, our MADS consistently outperforms the SOTA with comparable computation costs.

\vspace{-5pt}

\subsection{Qualitative Analysis}
\label{sec: qualitative_analysis}
\textbf{Visualization of Processes in Document Collection}.
In Figure~\ref{fig: MADS_document_collect}, we visualize processes of dividing documents by visual attributes ($p_{divide}$) and enriching less-described attribute documents ($p_{enrich}$). After the process of $p_{divide}$, we see that noisy descriptions (shown in \textcolor{gray}{gray}) in the collected document are removed in the divided document, such as descriptions about diet (the second last gray sentence) and lifestyles (the last gray sentence). Moreover, visual sentences in collected documents are divided into multiple paragraphs, where each paragraph only contains sentences about one attribute view (shown in the same color). However, some attribute views are not described in the collected document, \textit{i.e.}, \texttt{Fur and Texture}. After the process of $p_{enrich}$, each attribute paragraph contains rich visual descriptions for subsequent training.

\textbf{Visualization of Interpretable Predictions and Semantic Alignment}.
In Figure~\ref{fig: MADS_Improve}, we show the interpretable scores for the correct class and easily misclassified class in the left area (blue shaded) and salient image regions and corresponding top 5 attended words in the right area (orange shaded). We compute the cosine similarity between each single-view core feature and global image feature and offer interpretable scores for prediction. Compared to the I2MVFormer \cite{I2MVFormer}, our MADS gives the classification judgment based on multiple attribute views and predicts the correct category. For example, although sheep and horses have similar fur and habitats, our MADS predict correctly based on shape, color, and aggregate information. In addition, we observe that our MADS achieves accurate semantic alignment between visual words of each attribute view and corresponding image regions. This verifies the effectiveness of our MADS in knowledge transfer.

{
\begin{figure}[t]
\begin{center}
\centerline{\includegraphics[width=0.98\linewidth]{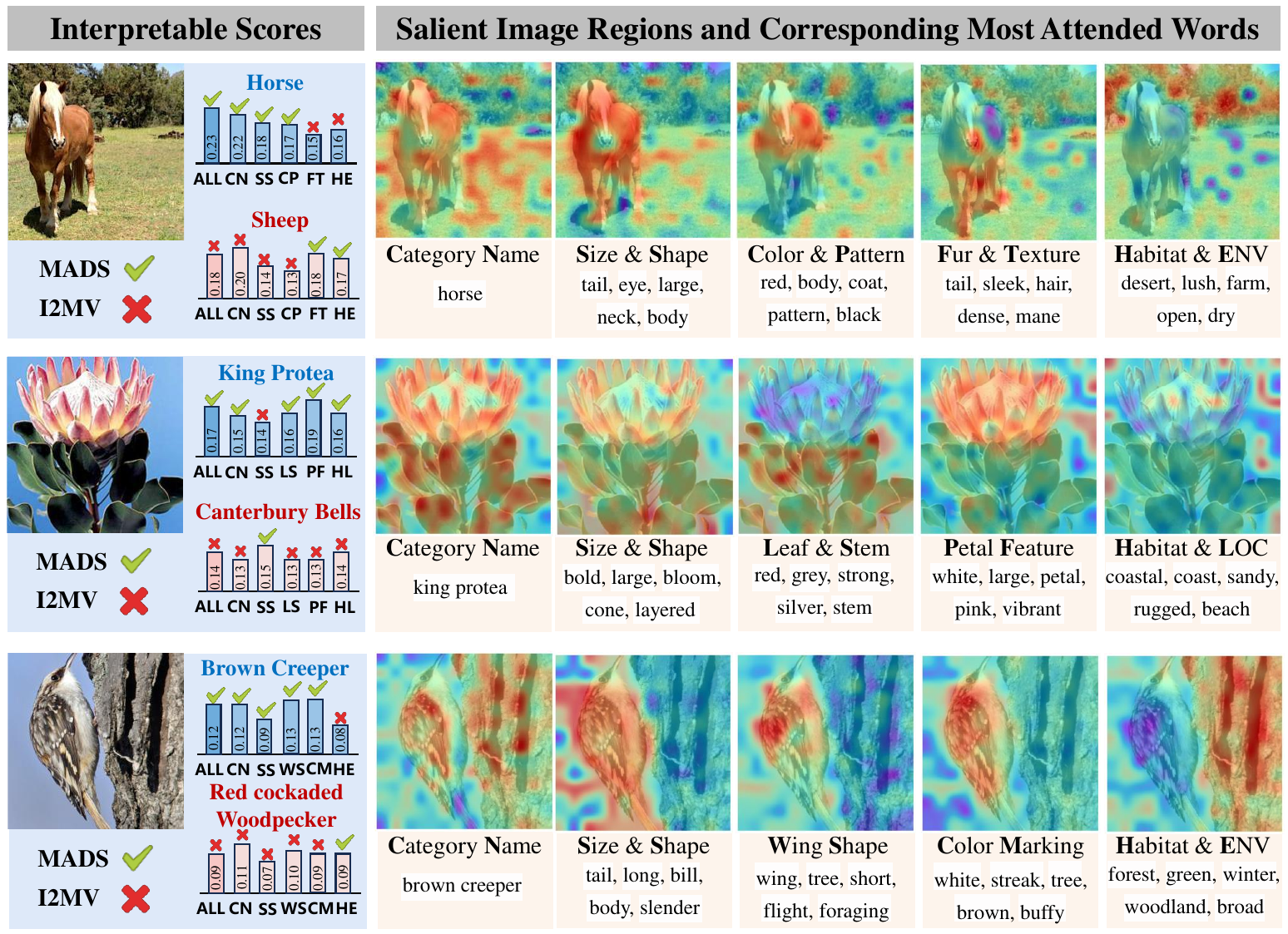}}
\vskip -0.07in
\caption{
Visualization of interpretable scores, salient regions, and attended words. 
Our MADS provides accurate and interpretable predictions from multiple views. Moreover, we see the semantic alignment between visual words (shown in the white area) of each attribute view and salient regions.
}
\label{fig: MADS_Improve}
\end{center}
\vskip -0.25in
\end{figure}
}




\section{Conclusion}
We propose a multi-attribute document supervision framework for document-based ZSL to filter noises in document collection and model training.
With the help of LLMs, we collect multi-attribute documents to remove non-visual descriptions and decouple semantic information before model training. Our MADS network extracts multi-view transferable knowledge for local and global semantic alignment. Besides, we introduce a focus loss to explicitly enhance the attention on visual words, also improving the performance of previous methods. Extensive experiments on public datasets validate the effectiveness of our MADS. Performance gains are qualitatively explained by accurate semantic alignment and interpretable scores from different views. Notably, compared to CLIP-based description methods, we show superior generalization under the situation without prior information on class names.




\bibliographystyle{IEEEtran}
\bibliography{reference}

\appendix

The supplementary material provides: 
\begin{itemize}
    \item Section~\ref{supp_sec: attribute_views}: Attribute views for each dataset.
    \item Section~\ref{supp_sec: ablation_views}: Ablation over attribute views in documents.
    \item Section~\ref{supp_sec: ablation_gpt4_version}: Ablation over versions of GPT-4 model.
    \item Section~\ref{supp_sec: compute_LLM_cost}: Computation costs for collecting documents.
    \item Section~\ref{supp_sec: add_qualitative}: Additional qualitative results.
    \item Section~\ref{supp_sec: train_details}: Training details.
    \item Section~\ref{supp_sec: visual_words}: Examples of visual words in document.
    \item Section~\ref{supp_sec: MAD_doc}: Examples of multi-attribute documents.
\end{itemize}

\subsection{Attribute Views for Each Dataset.}
\label{supp_sec: attribute_views}
In this section, we show LLM-generated visual attributes for each dataset, \textit{i.e.}, AWA2~\cite{AWA2} in Table~\ref{tab: visual_attributes_on_AWA2}, CUB~\cite{CUB_and_attribute_provide} in Table~\ref{tab: visual_attributes_on_CUB}, and FLO~\cite{FLO} in Table~\ref{tab: visual_attributes_on_FLO}. Notably, we see that all attribute views are visually relevant, which provides the basis for noise suppression and semantic decoupling.

{
    \begin{table}[htbp]
    \centering
    \caption{\textbf{Visual Attribute Views on AWA2 Dataset.}}
    \label{tab: visual_attributes_on_AWA2}
        \resizebox{0.99\linewidth}{!}
        {
        \begin{tabular}{p{2.1cm}|p{6cm}} 
        
        \toprule
        \textbf{Visual Attributes} &  \textbf{Explanation}\\
        \midrule
        
        Size and Shape & The overall body size and shape, including head, body, limbs, and tail, are primary indicators of species. \\ 
        \midrule
        Color and Patterns & Attention to coloration, including fur, skin, feather patterns, and unique markings like stripes or spots. \\ 
        \midrule
        Physical Features & Specific anatomical details such as the shape of beak, ears, nose, tail, paws, horns, or antlers, as well as eye color and shape, are significant. \\ 
        \midrule
        Fur, Feathers, or Scales Texture & The type and texture of body covering, whether it's fur, feathers, or scales, help in determining the species. \\ 
        \midrule
        Habitat and Environment & The animal's environment or habitat, evident from the image's background, can provide vital clues. \\ 
        
        \bottomrule
        \end{tabular}
        }
    \end{table}
}

{
    \begin{table}[htbp]
    \centering
    \caption{\textbf{Visual Attribute Views on CUB Dataset.}}
    \label{tab: visual_attributes_on_CUB}
        \resizebox{0.99\linewidth}{!}
        {
        \begin{tabular}{p{2.4cm} | p{6.5cm}} 
        \toprule
        \textbf{Visual Attributes} &  \textbf{Explanation}\\
        \midrule
        Size and Shape & The overall size (small, medium, large) and body shape of the bird. Different families of birds have distinctive shapes. \\
        \midrule
        Beak Shape and Size & The size and shape of the bird's beak can give clues about its diet and, consequently, its species. \\
        \midrule
        Color and Markings & The color patterns, including any distinctive markings, stripes, or spots. The colors of the head, back, underparts, and wings are particularly important. \\
        \midrule
        Legs and Feet & The length and color of the legs and the type of feet. \\
        \midrule
        Tail & The shape and length of the tail can be distinctive. \\
        \midrule
        Wing Shape & Shape and size of the wings, especially in flight. \\
        \midrule
        Behavior & Note distinctive behaviors, such as the way it flies, forages, or interacts with other birds. \\
        \midrule
        Habitat & The environment where the bird is found (e.g., woodland, wetland, grassland). \\
        \bottomrule
        \end{tabular}
        }
    \end{table}
}

{
    \begin{table}[htbp]
    \centering
    \caption{\textbf{Visual Attribute Views on FLO Dataset.}}
    \label{tab: visual_attributes_on_FLO}
        \resizebox{1.0\linewidth}{!}
        {
        \begin{tabular}{p{2.4cm}|p{6cm}} 
        \toprule
        \textbf{Visual Attributes} &  \textbf{Explanation}\\
        \midrule

        Petal Characteristics & This includes color and patterns on the petals, their shape (such as round, elongated, spiky), size, texture, and arrangement (overlapping, spaced, in single or multiple layers). \\
        \midrule

        Center of the Flower & This category encompasses the stamen and pistil, focusing on their color, structure, and any distinct features. This also includes the appearance of pollen and stamens. \\
        \midrule

        Leaf and Stem Features & The shape, size, color, arrangement, and pattern of leaves are vital. Stem characteristics, such as length, thickness, color, texture, and presence of hairs, are also included. \\
        \midrule

        Size and Shape of the Flower & This includes the overall size of the flower, both individual petals and total diameter, and the general shape. \\
        \midrule

        Patterns and Markings & Unique patterns or markings on the petals or leaves aid in identification. \\
        \midrule

        Habitat and Location & The environment where the flower is growing (like garden, forest, desert) and its geographical location. \\
         
        \bottomrule
        \end{tabular}
        }
    \end{table}
}

\subsection{Ablation over Attribute Views}
\label{supp_sec: ablation_views}
In this section, we ablate each attribute view for each dataset, \textit{i.e.}, AWA2~\cite{AWA2} in Table~\ref{tab: Ablation_views_AWA2}, CUB~\cite{CUB_and_attribute_provide} in Table~\ref{tab: Ablation_views_CUB}, and FLO~\cite{FLO} in Table~\ref{tab: Ablation_views_FLO}. We see that removing any attribute view will reduce performance, which verifies the effectiveness of each attribute view. Moreover, category names can also be considered an attribute view, which also plays a vital role (verified in Table IV of the main paper).

{
    \begin{table}[htbp]
    \centering
    \caption{\textbf{Ablation over Attribute Views on AWA2 Dataset.}}
    \label{tab: Ablation_views_AWA2}
        {
        \begin{tabular}{l | c | ccc}
        \toprule
        \multirow{2}{*}{\textbf{Model}} &  \textbf{AWA2} & \multicolumn{3}{c}{\textbf{AWA2}}\\
        \cmidrule(lr){2-2} \cmidrule(lr){3-5} 
        & \textbf{T1} & \textbf{U} & \textbf{S} & \textbf{H}\\
        \midrule
         \highlight{\textbf{MADS} (Baseline)} & \highlight{\textbf{86.7}} & \highlight{{81.5}} & \highlight{\textbf{86.3}} & \highlight{\textbf{83.9}} \\
         w/o Size and Shape & 84.2 & 77.0 & 84.5 & 80.6 \\
         w/o Color and Patterns & 83.1 & 80.4 & 82.4 & 81.4 \\
         w/o Physical Features & 81.8 & 78.4 & 83.5 & 80.9 \\
         w/o Fur, Feathers, or Scales Texture & 83.2 & \textbf{81.7} & 82.6 & 82.1 \\
         w/o Habitat and Environment & 79.9 & 78.5 & 81.9 &  80.2 \\
        \bottomrule
        \end{tabular}
        }
    \end{table}
}

{
    \begin{table}[ht]
    \centering
    \caption{\textbf{Ablation over Attribute Views on CUB Dataset.}}
    \label{tab: Ablation_views_CUB}
        {
        \begin{tabular}{l | c | ccc}
        \toprule
        \multirow{2}{*}{\textbf{Model}} &  \textbf{CUB} & \multicolumn{3}{c}{\textbf{CUB}}\\
        \cmidrule(lr){2-2} \cmidrule(lr){3-5} 
        & \textbf{T1} & \textbf{U} & \textbf{S} & \textbf{H}\\
        \midrule
         \highlight{\textbf{MADS} (Baseline)} & \highlight{\textbf{57.6}} & \highlight{\textbf{47.9}} & \highlight{{71.0}} & \highlight{\textbf{57.2}} \\
         w/o Size and Shape & 56.6 & 45.5 & 64.3 & 53.3 \\
         w/o Beak Shape and Size & 56.4 & 46.2 & 70.6 & 55.8 \\
         w/o Color and Markings & 57.3 & 46.6 & {72.8} & 56.9 \\
         w/o Legs and Feet & 57.4 & 46.6 & 68.2 & 55.3 \\
         w/o Tail & 56.4 & 46.4 & \textbf{73.5} & 56.9 \\
         w/o Wing Shape & 55.5 & 46.5 & 68.8 & 55.5 \\
         w/o Behavior & 56.1 & 46.8 & 65.9 & 54.8 \\
         w/o Habitat & 57.5 & 47.2 & 67.5 & 55.5 \\        
        \bottomrule
        \end{tabular}
        }
    \end{table}
}

{
    
    \begin{table}[ht]
    \centering
    \caption{\textbf{Ablation over Attribute Views on FLO Dataset.}}
    \label{tab: Ablation_views_FLO}
        {
        \begin{tabular}{l | c | ccc}
        \toprule
        \multirow{2}{*}{\textbf{Model}} &  \textbf{FLO} & \multicolumn{3}{c}{\textbf{FLO}}\\
        \cmidrule(lr){2-2} \cmidrule(lr){3-5} 
        & \textbf{T1} & \textbf{U} & \textbf{S} & \textbf{H}\\
        \midrule
         \highlight{\textbf{MADS} (Baseline)} & \highlight{\textbf{54.3}} & \highlight{\textbf{52.0}} & \highlight{\textbf{97.2}} & \highlight{\textbf{67.8}} \\
         w/o Petal Characteristics & 49.5 & 46.0 & 92.1 & 61.4 \\
         w/o Center of the Flower & 49.7 & 47.3 & 90.8 & 62.2 \\
         w/o Leaf and Stem Features & 48.7 & 44.8 & 91.5 & 60.1 \\
         w/o Habitat and Location & 47.0 & 44.5 & 97.1 & 61.0 \\
         w/o Size and Shape of the Flower & 46.7 & 43.3 & 89.5 & 58.3 \\
         w/o Patterns and Markings & 47.3 & 43.0 & 96.9 & 59.5 \\
        \bottomrule
        \end{tabular}
        }
    \end{table}
}

\subsection{Ablation over Different Versions of GPT-4 Model}
\label{supp_sec: ablation_gpt4_version}
In Table~\ref{tab: Ablation_on_GPT4_version}, we compare the performance of different versions of the GPT-4 model. 
The latest version achieves the best performance. Moreover, all results consistently outperform previous methods, further verifying our MADS's effectiveness.

{
    \begin{table}[!h]
    \centering
    \caption{\textbf{Ablation over Different Versions of GPT-4 Model.}}
    \label{tab: Ablation_on_GPT4_version}
        {
        \begin{tabular}{l ccc ccc}
        \toprule
        \multirow{2}{*}{\textbf{Model}}  
        & \multicolumn{3}{c}{\textbf{CUB}} & \multicolumn{3}{c}{\textbf{FLO}}\\
        \cmidrule(lr){2-4} \cmidrule(lr){5-7} 
        & \textbf{U} & \textbf{S} & \textbf{H} & \textbf{U} &  \textbf{S} & \textbf{H} \\
        \midrule
         GPT-4 (0613)  & 46.1 & 69.7  & 55.5 & 51.9  & 94.2 & 66.9 \\
         GPT-4 (1106)  & \textbf{47.9} & \textbf{71.0} & \textbf{57.6}& \textbf{52.0} & \textbf{97.2} & \textbf{67.8} \\
        \bottomrule
        \end{tabular}
        }
        \vskip 0.1in
    \end{table}
}

\subsection{Computation Cost Analysis for Collecting Documents}
\label{supp_sec: compute_LLM_cost}

In Table~\ref{tab: time_for_collect}, we show detailed time using LLMs for collecting multi-attribute documents for one category. Since we only generate visual views once, we average the time on category classes as the final time. 
Notably, we still save considerable human resources compared to annotating attributes.

{
    
    \begin{table}[!h]
    \centering
    \caption{\textbf{Computation Cost Analysis for Collecting Documents.}}
    \label{tab: time_for_collect}
        {
        \begin{tabular}{l c}
        \toprule
        \textbf{Process} & \textbf{Time}(s) \\
         
        \midrule
        generate visual attribute views (average) & 1 \\
        divide documents by visual attributes & 34 \\
        enrich less-described attribute documents & 37 \\
        \midrule
        \textbf{Total} & 72 \\
         \bottomrule
        \end{tabular}
        }
    \end{table}
}

\subsection{Additional Qualitative Results}
\label{supp_sec: add_qualitative}

\textbf{Baseline Performance Improvement.}
In Figure~\ref{fig:additional_base_improve}, we see that the model, after adding the $\mathcal{L}_{focus}$, accurately focuses on visual words and aligns visual words with corresponding image regions. For the baseline, the model incorrectly aligns non-visual words to image regions, which harms knowledge transfer and limits performance.

{
\begin{figure*}[htbp]
\begin{center}
\centerline{\includegraphics[width=0.85\linewidth]{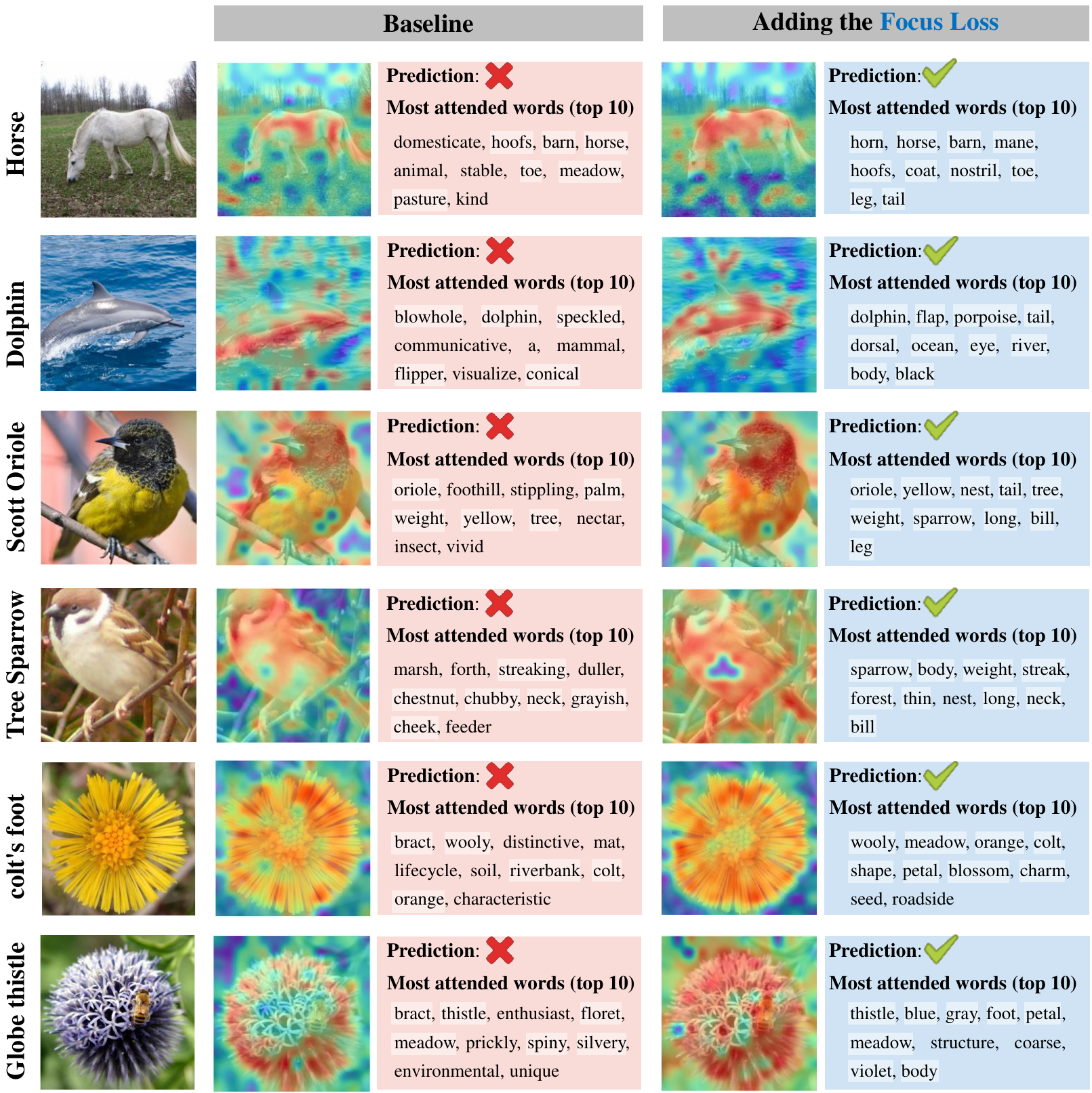}}
\caption{Visualization of salient image regions and most attended words in attention mechanism.
Although both the baseline and the model with the focus loss extract the salient visual regions, the latter pays more attention to visual words (shown with the white area) that are helpful for ZSL tasks.}
\label{fig:additional_base_improve}
\end{center}
\end{figure*}
}

\textbf{Performance Improvement Analysis.}
In Figure~\ref{fig:additional_MADS_improve}, MADS offers the correct prediction based on multiple views compared to I2MVFormer\cite{I2MVFormer}. Meanwhile, the performance gains are explained by accurate semantic alignment between each attribute view and corresponding image regions.

{
\begin{figure*}[htbp]
\begin{center}
\centerline{\includegraphics[width=0.85\linewidth]{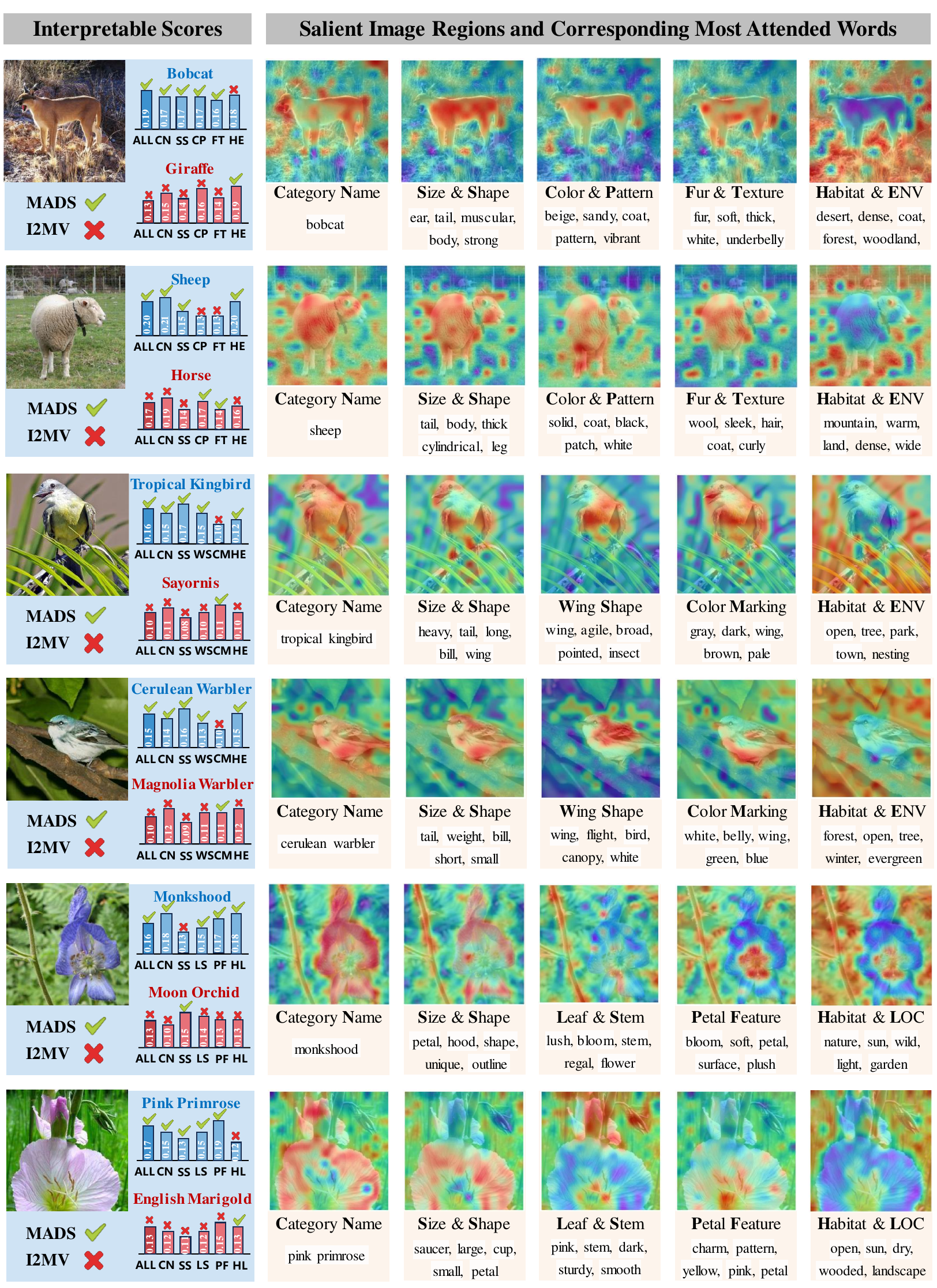}}
\caption{Visualization of interpretable scores, salient regions, and attended words. 
Our MADS provides accurate and interpretable predictions from multiple views. Moreover, we see the semantic alignment between visual words (shown in the white area) of each attribute view and salient regions.}
\label{fig:additional_MADS_improve}
\end{center}
\end{figure*}
}

\subsection{Training Details}
\label{supp_sec: train_details}

MADS is implemented in PyTorch and trained with a Nvidia GeForce RTX 3090 GPU.
We use the ViT-B/16 \cite{vit} pre-trained on ImageNet 1K~\cite{ImageNet} as the visual backbone, which respects the GUB split~\cite{AWA2}.
This means no unseen classes are used to train the visual backbone.
The detailed hyperparameters are shown in Table~\ref{tab:train_details} for three datasets.

    \begin{table*}[htbp]
    \centering
    \caption{\textbf{Hyperparameters Settings for AWA2, CUB, and FLO Datasets.}}
    \label{tab:train_details}
        {
        \begin{tabular}{l ccc}
        \toprule
        \textbf{Config} &  \textbf{AWA2}~\cite{AWA2} & \textbf{CUB}~\cite{CUB_and_attribute_provide} & \textbf{FLO}~\cite{FLO}
        \\
        \midrule
        \textbf{Regular Training Setting} & & & \\
        optimizer & AdamW\cite{Adamw} & AdamW\cite{Adamw} & AdamW\cite{Adamw}  \\
        base learning rate & 1.5e-4 & 1e-3 & 7e-4 \\
        dropout & 0.25 & 0.15 & 0.15 \\
        weight decay & 0.05 & 0.05 & 0.05 \\
        batch size & 64 & 64 & 64 \\
        learning rate schedule & cosine decay & cosine decay & cosine decay \\
        warmup epochs & 0 & 3 & 0 \\
        epochs & 40 & 40 & 40 \\
        augmentation & RandomResizedCrop & RandomResizedCrop & RandomResizedCrop \\ 
        \midrule
        \textbf{Specific Settings in MADS} & & & \\
        $\mathcal{\lambda}_{focus}$ & 0.5 & 0.5 & 0.5 \\
        $\mathcal{\lambda}_{local}$ & 0.2 & 0.5 & 0.5 \\
        fuse ratio $\beta$ & 0.5 & 0.5 & 0.5 \\
        number of semantic queries K & 4 & 4 & 8 \\ 
        LLM $\tau$ & 1.0 & 1.0 & 1.0 \\
        dimension of semantic embedding $r$ & 256 & 128 & 128 \\
        layers of text encoder & 2 & 2 & 2 \\
        layers of semantic perceiver & 2 & 2 & 2 \\
        layers of aggregation module & 1 & 1 & 1 \\
        \bottomrule
        \end{tabular}
        }
    \end{table*}

\subsection{Examples of Visual Words in Document}
\label{supp_sec: visual_words}

{
We show one example of visual words in documents for AWA2 in Figure~\ref{fig:visual_word_AWA2}, CUB in Figure~\ref{fig:visual_word_CUB}, and FLO in Figure~\ref{fig:visual_word_FLO}. Visual words are the adjectives and nouns that visually describe a class and offer a bridge to transfer knowledge from seen to unseen classes. Meanwhile, we also see that other words offer contextual information for visual words, which are also crucial for recognition.

{
\begin{figure}[t]
\begin{center}
\centerline{\includegraphics[width=1.0\linewidth]{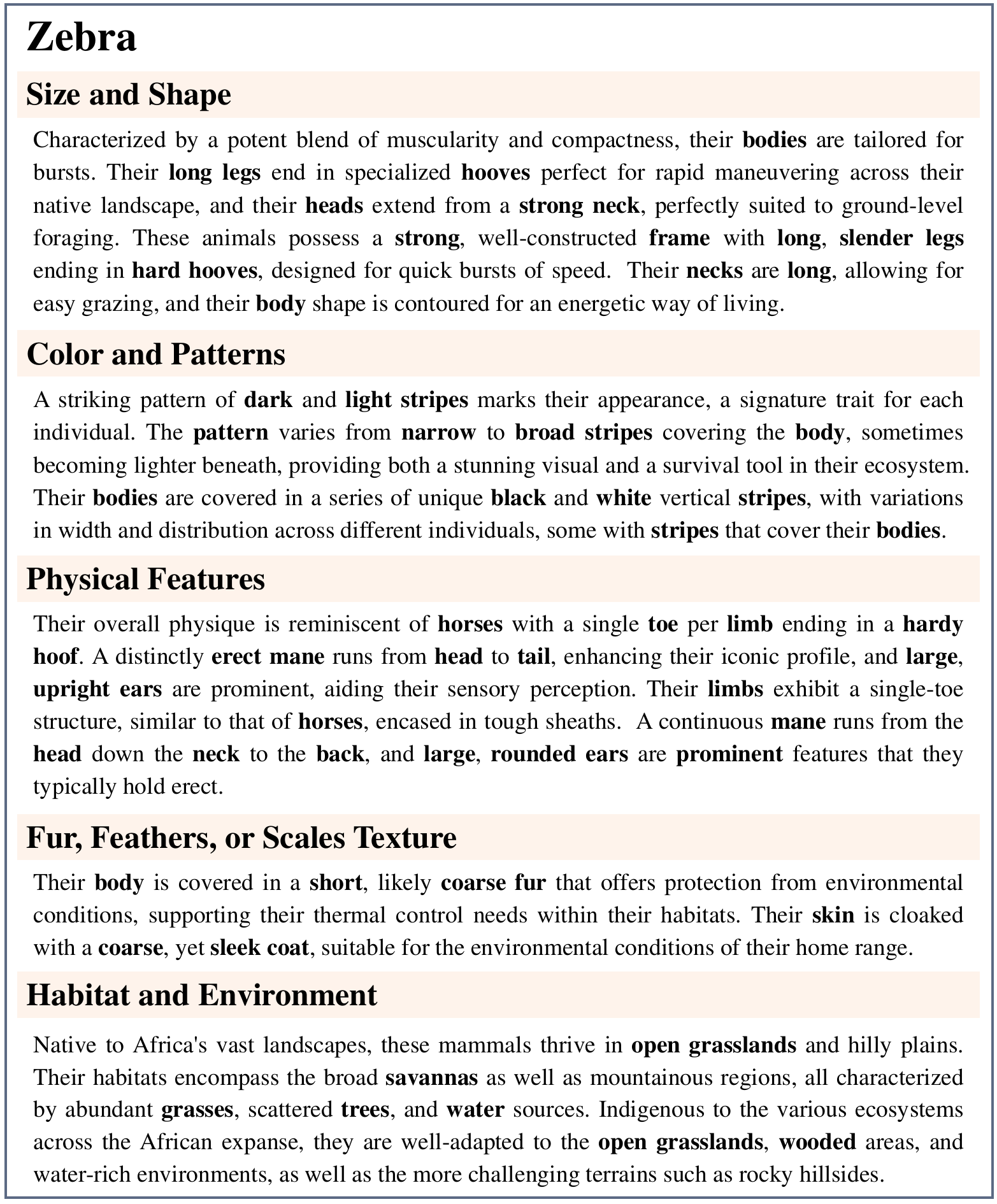}}
\caption{The example of visual words in documents for AWA2 dataset. 
The visual words are in \textbf{bold}.}
\label{fig:visual_word_AWA2}
\end{center}
\end{figure}
}

{
\begin{figure}[t]
\begin{center}
\centerline{\includegraphics[width=1.0\linewidth]{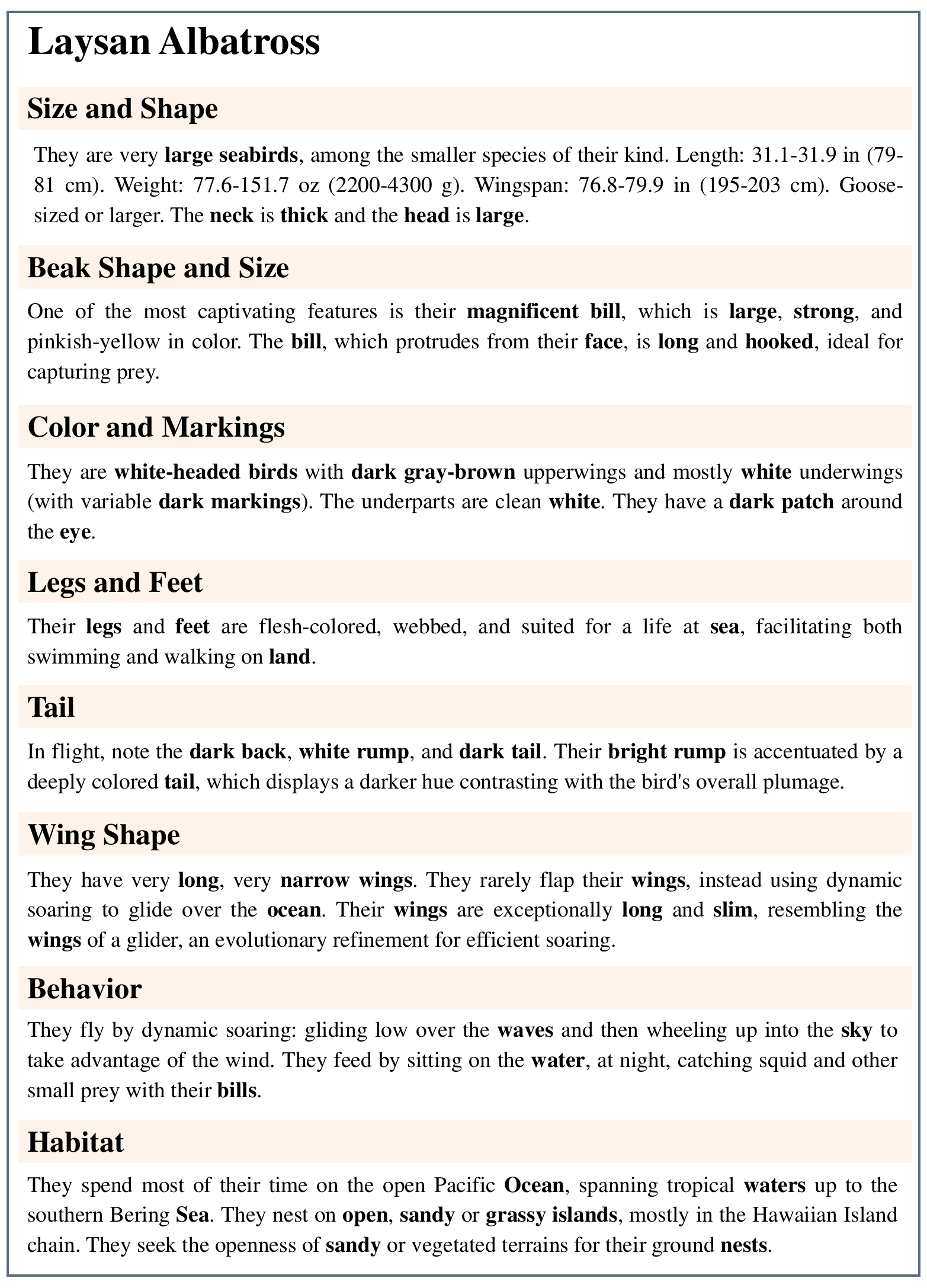}}
\caption{The example of visual words in documents for CUB dataset. 
The visual words are in \textbf{bold}.}
\label{fig:visual_word_CUB}
\end{center}
\end{figure}
}

{
\begin{figure}[t]
\begin{center}
\centerline{\includegraphics[width=1.0\linewidth]{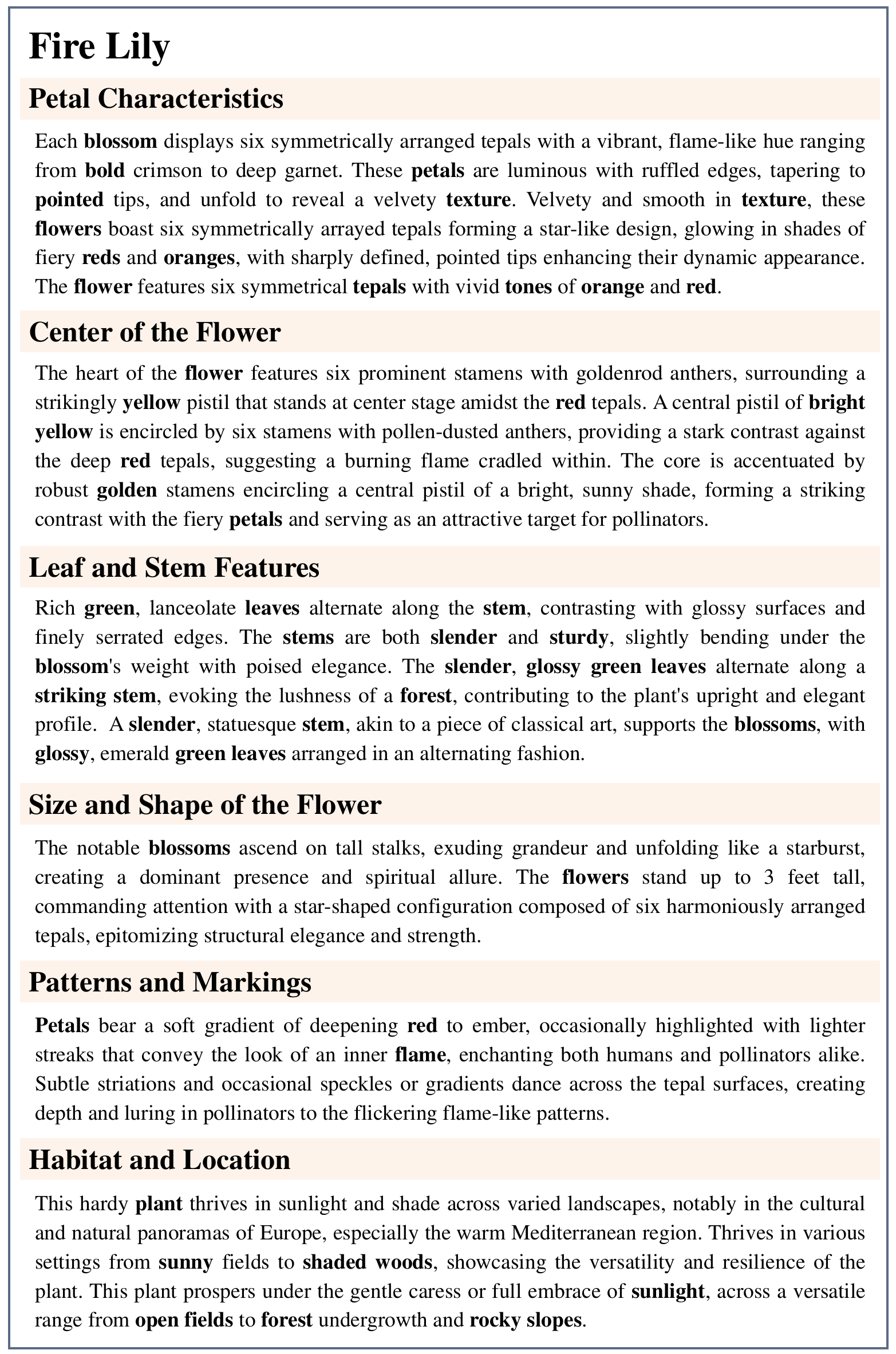}}
\caption{The example of visual words in documents for FLO dataset. 
The visual words are in \textbf{bold}.}
\label{fig:visual_word_FLO}
\end{center}
\end{figure}
}

}


\subsection{Examples of Collected Multi-Attribute Document}
\label{supp_sec: MAD_doc}

We show two examples of multi-attribute documents for three datasets as follows.

\noindent \textbf{AWA2: Giraffe.}
\begin{itemize}
    \item \textbf{Size and Shape}: Characterized by an extraordinary vertical silhouette with notably elongated limbs that support their towering form. A gracefully curved neck extends upward and merges with a compact body. The legs are svelte but sturdy, a testament to elegant strength, and the tail, featuring a decorative tuft, punctuates their distinct physical profile. Exhibiting a unique profile, these animals have extraordinarily long necks and legs, disproportionate to their compact bodies, and are finished with a tufted tail for balance.  Their elevated height is a byproduct of extreme evolutionary development. 
    \item \textbf{Color and Patterns}: A dynamic assortment of variegated patches adorns their coat in a spectrum of earthy tones, from sandy to deep chestnut, arrayed on a cream base. These naturally tessellated markings are unique to each individual, much like a biological fingerprint. Their pelts display a pattern of variegated spots that range from deep brown to reddish hues on a lighter backdrop, each individual sporting a unique array of these intricate markings that provide both concealment and temperature control.
    \item \textbf{Physical Features}: Possessing large, expressive eyes positioned on a small face, they maintain a wide range of vision. Atop their head are modest, horn-like protrusions, enveloped in skin, while their neatly edged ears are in proportion with their delicate auditory necessities. Equipped with large, dark eyes granting an extensive field of view, modest protrusions atop their head for combat, and ears finely tuned to the environment, these animals possess keen senses to monitor their expansive territories. 
    \item \textbf{Fur, Feathers, or Scales Texture}: Their skin is cloaked in a velvety smooth, yet occasionally bristly, pelt that serves both as protection against the elements and as a natural disguise within their habitat. Their coat consists of patch-specific fur textures, with the main body covered in shorter, coarser hair, and a softer underside, while the tail ends in a dense swishing fringe to ward off insects.
    \item \textbf{Habitat and Environment}: Inhabitants of expansive savannas and woodlands, these creatures take advantage of their considerable height for both predator evasion and foraging. The open vistas of these environments are ideal for their lifestyle and nutritional needs. Native to expansive African terrains, they roam areas populated by sparse forests and open plains, utilizing their considerable height to forage foliage well above ground level, and influencing the botanical diversity of their ecosystem. 
\end{itemize}

\noindent \textbf{AWA2: Horse.}
\begin{itemize}
    \item \textbf{Size and Shape}: Magnificently built with a range from muscular and robust to slim and elegant. They possess long necks that gracefully support their large heads, complete with wide foreheads and expressive, large eyes. Their ears are notably mobile and adept at picking up nuanced sounds. A notable mane flows down the neck, adding grandeur and minimal protection, while a proportionately shorter, thickly-haired tail acts as an insect repellent. These animals are large and either muscular or svelte, boasting long necks and proportionate heads that provide a wide field of vision.  Their bodies are engineered for a balance of speed and strength, a characteristic that is well-suited to their varied lifestyles.
    \item \textbf{Color and Patterns}: Their coat exhibits a spectrum of colors from solid tones to complex spottings and rosettes, especially in those with a heritage marked by a myriad of spots. Some exhibit a patchwork of colors across their body, creating a dappled effect, while certain individuals have hooves strikingly patterned in black and white, resembling piano keys. Their coats come in a spectrum of colors ranging from deep blacks to radiant reds and can be uniform or feature patterns like piebald, skewbald, pinto, and palomino.  Such diversity in coloration is a testament to their extensive breeding history and adaptations.
    \item \textbf{Physical Features}: The definition of stature is reflected in the sturdy hooves at the end of each limb, which carry their considerable weight with ease. The hooves, typically encased in a hard, keratinous material, offer endurance across various landscapes. While predominantly black, some hooves may reflect the color of the legs. Their form is powerfully structured, with strong hindquarters and ample chest, underpinning their athletic abilities. Equipped with single-toe limbs capped by durable hooves, they are designed for efficient locomotion.  Their robust chests accommodate large lungs, and their specialized teeth are adapted for a herbivorous diet, capable of grinding and cropping vegetation.
    \item \textbf{Fur, Feathers, or Scales Texture}: Their coat is velvety, especially around the muzzle, becoming denser on the body, and is especially radiant when cared for. The mane and tail feature coarser hair strands, capable of elegant motion when airborne. This range in hair texture reflects their ability to thrive in diverse environmental conditions. Their bodies are covered in a coat that varies from dense and fluffy to sleek and fine depending on the season.  They possess coarse manes and tails, which can be indicators of grooming and are utilized for pest control and communication.
    \item \textbf{Habitat and Environment}: These robust animals thrive in an array of environments from dry to lush landscapes. In the wild, they prefer open areas like steppes and prairies, but have also become accustomed to domesticated settings, such as farms or training facilities. They flourish in spaces that accommodate their need for active movement. They are versatile in their environmental occupancy, inhabiting a range of climates from arid deserts to chilly Arctic fringes.  They tend to favor open areas where they can freely move and forage, with essential access to water and food sources.
\end{itemize}

\textbf{CUB: Brown Creeper.}
\begin{itemize}
    \item \textbf{Size and Shape}: Tiny, lanky songbirds with long, spine-tipped tails, slim bodies, and slender, decurved bills. Length: 4.7-5.5 in (12-14 cm). Weight: 0.2-0.3 oz (5-10 g). Wingspan: 6.7-7.9 in (17-20 cm).
    \item \textbf{Beak Shape and Size}: The bill is slender and decurved, perfect for probing into crevices in tree bark to find insects and spiders.
    \item \textbf{Color and Markings}: They have streaked brown and buff upperparts, with a broad, buffy stripe over the eye. The underparts are white, usually hidden against the tree trunk.
    \item \textbf{Legs and Feet}: Their legs and feet are specialized for clinging to tree trunks, supporting their unique foraging behavior.
    \item \textbf{Tail}: The tail is long and spine-tipped, used for support as they hitch upward in a spiral around tree trunks.
    \item \textbf{Wing Shape}: Their wings are well-suited for short flights between trees, necessary for their foraging style.
    \item \textbf{Behavior}: They forage by hitching upward in a spiral around tree trunks and limbs, using their stiff tails for support, and fly weakly to the base of another tree to continue foraging.
    \item \textbf{Habitat}: They are found in mature evergreen or mixed evergreen-deciduous forests for breeding. In winter, they can be found in a broader variety of forests, including deciduous woodlands.
\end{itemize}

\noindent \textbf{CUB: Mockingbird.}
\begin{itemize}
    \item \textbf{Size and Shape}: Medium-sized songbird, more slender than a thrush with a longer tail. Length: 8.3-10.2 in (21-26 cm). Weight: 1.6-2.0 oz (45-58 g). Wingspan: 12.2-13.8 in (31-35 cm).
    \item \textbf{Beak Shape and Size}: Long, thin bill with a hint of a downward curve.
    \item \textbf{Color and Markings}: Overall gray-brown, paler on the breast and belly. Two white wingbars on each wing and a white patch in each wing. White outer tail feathers are also flashy in flight.
    \item \textbf{Legs and Feet}: Long legs that are well-adapted for running and hopping on the ground.
    \item \textbf{Tail}: Long tail that is gray-brown like the body, which appears particularly long in flight and aids in balance and maneuverability.
    \item \textbf{Wing Shape}: Short, rounded, and broad wings, which are efficient for quick takeoffs and agile flight.
    \item \textbf{Behavior}: They are known for their songs and mimicry. They sit conspicuously on high vegetation, fences, eaves, or telephone wires, or run and hop along the ground. They are territorial and will aggressively chase off intruders.
    \item \textbf{Habitat}: Found in towns, suburbs, backyards, parks, forest edges, and open land at low elevations.
\end{itemize}

\textbf{FLO: King Protea.}
\begin{itemize} 
    \item \textbf{Petal Characteristics}: Sizable, ranging from 4 to 6 inches, these petals showcase a splendid spectrum from creamy hues and soft pinks to vibrant crimsons and corals, with a sumptuous velvety texture occasionally highlighted by a waxy sheen. Thick and velvety, the petals are wide and come in a pallete ranging from pastels to deep, vivid colors, with a luxurious sheen highlighting their texture.  Commanding blossoms formed by robust, hand-sized petals display an exquisite color range from soft whites and pinks to deep reds and oranges.  Their suede-like surface is both visually splendid and durable, with a waxy sheen.
    \item \textbf{Center of the Flower}: A prominent cone at the core is encircled by a spray of feathery styles, creating depth and complexity while drawing attention to the detailed structure. The flower's core is dominated by a prominent, cone-like structure, adorned with a profusion of feathery, needle-like projections that add an exotic texture reminiscent of a regal crown.  The bloom's core is a striking cluster of plumose styles, creating a texture similar to luxurious furled feathers, enhancing the flower's dramatic and commanding presence.
    \item \textbf{Leaf and Stem Features}: A strong stem supports the bloom, topped by a ring of sturdy, leaf-like bracts that vary in color from tranquil greens and silvery greys to deep earthen reds, offering both aesthetic appeal and protection. A circle of tough, leaf-like bracts cradles the flower, providing support and defense in varying shades of green, silver, and red, enhancing the flower's bold silhouette.  At the bloom's base, an array of sturdy, sea anemone-like bracts in muted greens, steely greys, and splashes of vermillion robustly encircles the flower, reminiscent of a protective, ornate collar. 
    \item \textbf{Habitat and Location}: Native to South Africa's southwestern coastal regions, it thrives in diverse environments including sandy beaches, rugged mountains, and arid areas, contributing to the region's ecological variety. Native to diverse South African landscapes, this bloom thrives across sandy coasts, mountain slopes, and arid regions, indicative of its fynbos biome origin.  Native to the Western Cape's fynbos ecosystem, these plants thrive in sandy shores, mountainous enclaves, and arid plains, symbolizing their resilience and ecological significance. 
    \item \textbf{Size and Shape of the Flower}: A majestic, globe-shaped inflorescence that can reach an impressive width of up to 12 inches, consisting of multiple layers of petals spiraling around a bold, central cone. This bloom presents as a substantial, spherical cluster, with a notable central core and layered petals that collectively create an imposing, globe-like form.  The immense dome-shaped blooms feature a distinctive conical center, exuding royal grandeur and perfect geometrical beauty.
    \item \textbf{Patterns and Markings}: Occasionally the petals exhibit delicate striations or speckled details that enhance their vivid tones, adding depth and a sense of individuality to each bloom. Hints suggest the petals may feature subtle distinctive patterns such as speckles or striations, imparting unique visual textures to each flower.  Each flower features unique markings, such as subtle spots or striations, that create a personal fingerprint and contribute to its intricate visual story.
\end{itemize}

\noindent \textbf{FLO: Sunflower.}
\begin{itemize}
    \item \textbf{Petal Characteristics}: These flowers brandish ray florets, mimicking petals in shape, ranging from lanceolate to pointed tips. Their colors span a spectrum reminiscent of twilight, from radiant yellows to fiery oranges, rich russets, and on occasion, muted pastels and pure whites. Some boast striking bi-colored patterns that contrast the greenery beneath them, crafting a visual delight rooted in nature's palette. Elongated ray florets extend outward much like the sun's rays, featuring a spectrum from bright, sunny yellow to striking russet reds.  They possess a texture that is at once tender and resilient, with an aptitude for shining brightly and attracting the interest of various pollinators.  Bright and expressive, these blooms feature a halo of predominantly sunny yellow ray florets surrounding their faces, with possible fiery shades of red and orange or softer tones of cream and copper.  The florets exhibit a durable, velvety texture and are arranged symmetrically, exuding a strong presence.
    \item \textbf{Center of the Flower}: The floral centerpiece is a cluster of disc florets, each a tiny embodiment of symmetry, spiraling together to form a compact heart. Deep shades from brown to purple set the stage, surrounded by a vivid ring of ray florets. This dark tapestry of blooms signals its life cycle's progression by the subtle swelling of pollinated florets. At the core, numerous tiny disc florets cluster tightly, revealing a dense array of flowers crucial for reproduction.  These florets, ranging in color from deep browns to rich burgundy, present an eye-catching hub that not only serves as a visual focal point but also plays a vital role in pollination.  The heart of the bloom resembles a dark, intricate constellation, composed of many small tubular disc florets with exposed stigmas and stamens that beckon to pollinators.  Displaying a rich palette of browns and burgundies, the center harmonizes reproductive function with visual appeal.
    \item \textbf{Leaf and Stem Features}: These plants are supported by substantial leaves and stems. Leaves emerge broad and coarse, with a fine layer of hairs to conserve moisture. Stems, green with hues of purple, are stout, capable of rising to impressive heights, holding aloft the grandeur of the blooms. The large, coarse leaves range from ovate to heart-shaped, primarily deep green with prominent veining.  They are borne on thick, sturdy stems capable of supporting the weight of the bloom.  The stems are textured and strong, demonstrating the plant's overall robustness in both form and vigor.  The foliage is a striking green, in heart or lance shapes with serrated edges, contrasting the blooms.  The stems are rugged and hairy, providing the necessary strength to uphold the substantial blooms and reflecting the plant's overall resilience.
    \item \textbf{Habitat and Location}: Originating from the demanding terrains of North America, these plants are comfortable across various settings, from flat, open fields to rocky high-altitude terrains. Their adaptability is conspicuous; they flourish in nutrient-rich, water-abundant soils, transforming the landscape into a burst of radiant color. Originating in North America, these plants thrive under the full sun in a variety of landscapes, from fertile soil to arid conditions.  Their adaptability has allowed them to establish themselves globally, marked by their resilience and ability to prosper in diverse environments.  Native to sunny expanses and tolerant of diverse climatic conditions, these plants are at home on open plains and in various global locales.  They thrive best in rich soils but are adaptable to less fertile environments as well. 
    \item \textbf{Size and Shape of the Flower}: Their inflorescences mimic the sun, with grand corollas unfurling to create impressive displays that can range from the modest to the monumental. Whether a singular, striking bloom or a cluster of smaller heads, each variation lends its own unique appeal. With a nod to the sun, each magnificent floral head leaves a bold statement, varying in size but consistently noteworthy.  The considerable size aids the plant to excel in attracting pollinators, thereby enhancing its survival through increased opportunities for reproduction.  The flower heads are impressively large and predominantly circular, sometimes spanning over 30 centimeters.  They possess a radial arrangement of florets that resemble the sun's rays, offering both visual impact and structural complexity.
    \item \textbf{Patterns and Markings}: The dark-colored, densely packed disc florets form a complex, spiraling pattern that blooms from the outside in, offering a dynamic visual experience that complements the striking surround of colorful 'petals.' The core of these flowers is a masterclass in the beauty of collective intricacy. Upon close inspection, the central disc reveals a captivating pattern of smaller florets, together forming a visually complex structure that doubles as a beacon for pollinators.  These patterns, in hues of earth and dusk, contribute to a signature aesthetic that's enchanting to observers.  The central disc florets create a captivating, organic mosaic of spiraling patterns and muted yet intense colors, contrasting with the vibrant ring of ray florets and enhancing the flower's pollination efficiency.
\end{itemize}

\vfill

\end{document}